\documentclass[10pt,twocolumn,letterpaper]{article}

\usepackage{cvpr}      
\usepackage[accsupp]{axessibility}
\usepackage{amsmath}
\usepackage{multirow}
\usepackage[normalem]{ulem}
\usepackage{xcolor}
\usepackage{algorithm}
\usepackage{algpseudocode}
\usepackage{float} 
\usepackage{listings}

\newcommand\blfootnote[1]{
    \begingroup
    \renewcommand\thefootnote{}\footnote{#1}
    \addtocounter{footnote}{-1}
    \endgroup
}

\definecolor{cvprblue}{rgb}{0.21,0.49,0.74}
\usepackage[pagebackref,breaklinks,colorlinks,citecolor=cvprblue]{hyperref}
\usepackage{dsfont}
\def\paperID{7957} 
\def\confName{CVPR}
\def\confYear{2024}

\title{
DreamMatcher: Appearance Matching Self-Attention\\ for Semantically-Consistent Text-to-Image Personalization}

\begin{document}
\author {
    Jisu Nam$^*$\textsuperscript{\rm 1},
    Heesu Kim\textsuperscript{\rm 2},
    DongJae Lee\textsuperscript{\rm 2},
    Siyoon Jin\textsuperscript{\rm 1},
    Seungryong Kim$^\dag$\textsuperscript{\rm 1}, 
    Seunggyu Chang$^\dag$\textsuperscript{\rm 2} \\ \\
    \textsuperscript{\rm 1}Korea University \hspace{5pt}
    \textsuperscript{\rm 2}NAVER Cloud \\
    {\tt\small\ \href{https://ku-cvlab.github.io/DreamMatcher}{https://ku-cvlab.github.io/DreamMatcher}}
 \\
}


\newcommand{\paragrapht}[1]{\noindent\textbf{#1}}
\twocolumn[{
\renewcommand\twocolumn[1][]{#1}%
\maketitle
\begin{center}
\captionsetup{type=figure}
\vspace{-2pt}
\includegraphics[width=0.99\textwidth]{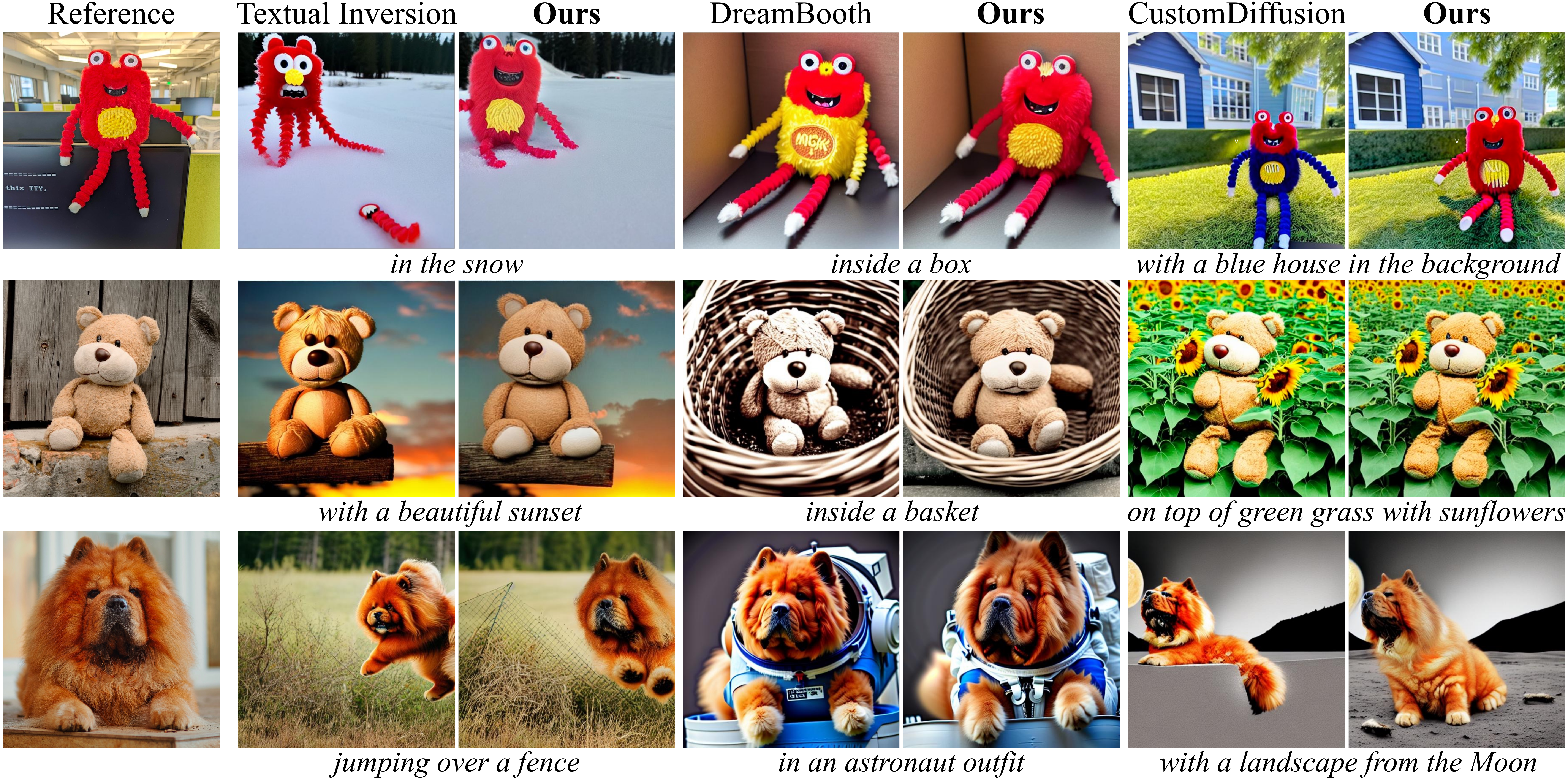}
\vspace{-8pt}
\captionof{figure}{\textbf{DreamMatcher enables semantically-consistent text-to-image (T2I) personalization.} Our DreamMatcher is designed to be compatible with any existing T2I personalization models, without requiring additional training or fine-tuning. When integrated with them, DreamMatcher significantly enhances subject appearance, including colors, textures, and shapes, while accurately preserving the target structure as guided by the target prompt.}
\label{qual:teaser}\vspace{-10pt}
\end{center}
}]
\maketitle{}

\blfootnote{$^\dag$Co-corresponding author.}
\blfootnote{$^*$Work done during an internship at NAVER Cloud.}

\begin{abstract}

{The objective of text-to-image (T2I) personalization is to customize a diffusion model to a user-provided reference concept, generating diverse images of the concept aligned with the target prompts. Conventional methods representing the reference concepts using unique text embeddings often fail to accurately mimic the appearance of the reference. To address this, one solution may be explicitly conditioning the reference images into the target denoising process, known as key-value replacement. However, prior works are constrained to local editing since they disrupt the structure path of the pre-trained T2I model. To overcome this, we propose a novel plug-in method, called DreamMatcher, which reformulates T2I personalization as semantic matching. Specifically, DreamMatcher replaces the target values with reference values aligned by semantic matching, while leaving the structure path unchanged to preserve the versatile capability of pre-trained T2I models for generating diverse structures. We also introduce a semantic-consistent masking strategy to isolate the personalized concept from irrelevant regions introduced by the target prompts. Compatible with existing T2I models, DreamMatcher shows significant improvements in complex scenarios. Intensive analyses demonstrate the effectiveness of our approach.}

\end{abstract}    
\section{Introduction}
\label{sec:intro}
The objective of text-to-image (T2I) personalization~\cite{gal2022image, ruiz2023dreambooth, kumari2023multi} is to customize T2I diffusion models based on the subject images provided by users. Given a few reference images, they can generate novel renditions of the subject across diverse scenes, poses, and viewpoints, guided by the target prompts.

Conventional approaches~\cite{gal2022image, voynov2023p+, ruiz2023dreambooth, kumari2023multi, han2023svdiff, dong2022dreamartist} for T2I personalization often represent subjects using unique text embeddings~\cite{radford2021learning}, by optimizing either the text embedding itself or the parameters of the diffusion model. However, as shown in Figure~\ref{qual:teaser}, they often fail to accurately mimic the appearance of subjects, such as colors, textures, and shapes. This is because the text embeddings lack sufficient spatial expressivity to represent the visual appearance of the subject~\cite{hao2023vico, radford2021learning}. To overcome this, recent works~\cite{jia2023taming, shi2023instantbooth, chen2023subject, xiao2023fastcomposer, gal2023designing, wei2023elite, chen2023photoverse, su2023identity, li2023blip} enhance the expressivity by training T2I models with large-scale datasets, but they require extensive text-image pairs for training.

To address the aforementioned challenges, one solution may be explicitly conditioning the reference images into the target denoising process. Recent subject-driven image editing techniques~\cite{cao2023masactrl, mou2023dragondiffusion, chen2023anydoor, chen2023fec, huang2023kv, khandelwal2023infusion} propose conditioning the reference image through the self-attention module of a denoising U-Net, which is often called key-value replacement. In the self-attention module~\cite{ho2020denoising}, image features from preceding layers are projected into queries, keys, and values. They are then self-aggregated by an attention operation~\cite{vaswani2017attention}. Leveraging this mechanism, previous image editing methods~\cite{cao2023masactrl, mou2023dragondiffusion} replace the keys and values from the target with those from the reference to condition the reference image into the target synthesis process. As noted in~\cite{tumanyan2023plug, tewel2023key, balaji2022ediffi, hertz2022prompt}, we analyze the self-attention module into two distinct paths having different roles for T2I personalization: the query-key similarities form the \textit{structure} path, determining the layout of the generated images, while the values form the \textit{appearance} path, infusing spatial appearance into the image layout. 

\begin{figure}[t]
    \centering
    \includegraphics[width=0.48\textwidth]{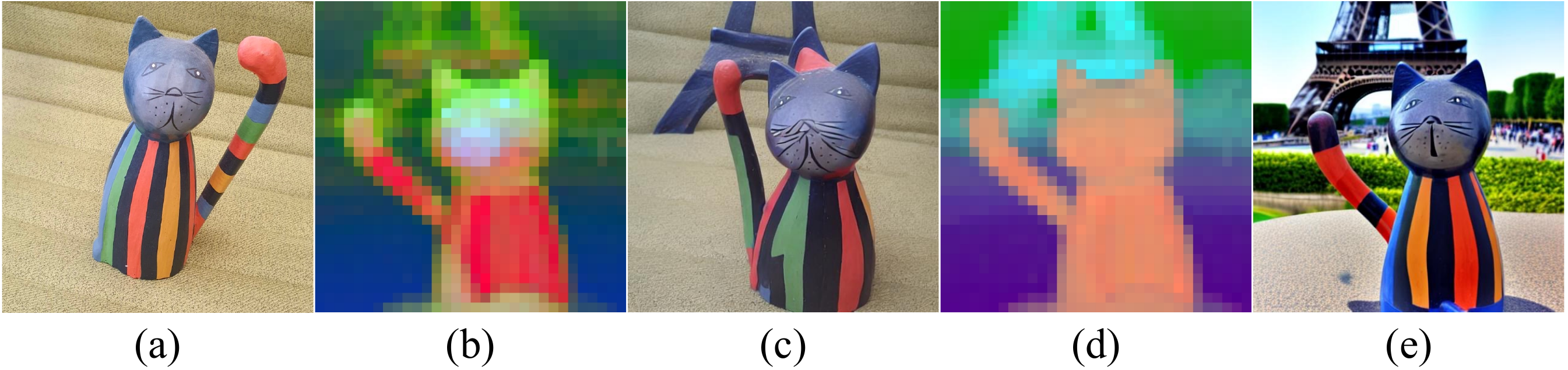}
    \vspace{-15pt}
    \caption{\textbf{Intuition of DreamMatcher:} (a) reference image, (b) disrupted target \textit{structure} path by key-value replacement~\cite{cao2023masactrl, mou2023dragondiffusion, chen2023anydoor, chen2023fec, huang2023kv, khandelwal2023infusion}, (c) generated image by (b), (d) target \textit{structure} path in pre-trained T2I model~\cite{ruiz2023dreambooth}, and (e) generated image by DreamMatcher. For visualization, principal component analysis (PCA)~\cite{pearson1901liii} is applied to the structure path. Key-value replacement disrupts the target structure, yielding sub-optimal personalized results, whereas DreamMatcher better preserves the target structure, producing high-fidelity subject images aligned with target prompts.}
    \vspace{-10pt}
\label{qual:motivation}
\end{figure}

As demonstrated in Figure~\ref{qual:motivation}, our key observation is that the replacement of target keys with reference keys in the self-attention module disrupts the structure path of the pre-trained T2I model. Specifically, an optimal key point for a query point can be unavailable in the replaced reference keys, leading to a sub-optimal matching between target queries and reference keys on the structure path. Consequently, reference appearance is then applied based on this imperfect correspondence. For this reason, prior methods incorporating key and value replacement often fail at generating personalized images with large structural differences, thus being limited to local editing. To resolve this, ViCo~\cite{hao2023vico} incorporates the tuning of a subset of model weights combined with key and value replacement. However, this approach necessitates a distinct tuning process prior to its actual usage.

In this paper, we propose a plug-in method dubbed \textbf{\texttt{DreamMatcher}} that effectively transfers reference appearance while generating diverse structures. DreamMatcher concentrates on the appearance path within the self-attention module for personalization, while leaving the structure path unchanged. However, a simple replacement of values from the target with those from the reference can lead to structure-appearance misalignment. To resolve this, we propose a matching-aware value injection leveraging semantic correspondence to align the reference appearance toward the target structure. Moreover, it is essential to isolate only the matched reference appearance to preserve other structural elements of the target, such as occluding objects or background variations. To this end, we introduce a semantic-consistent masking strategy, ensuring selective incorporation of semantically consistent reference appearances into the target structure. Combined, only the correctly aligned reference appearance is integrated into the target structure through the self-attention module at each time step. However, the estimated reference appearance in early diffusion time steps may lack the fine-grained subject details. To overcome this, we introduce a sampling guidance technique, named semantic matching guidance, to provide rich reference appearance in the middle of the target denoising process.

DreamMatcher is compatible with any existing T2I personalized models without any training or fine-tuning. We show the effectiveness of our method on three different baselines~\cite{gal2022image, ruiz2023dreambooth, kumari2023multi}. DreamMatcher achieves state-of-the-art performance compared with existing tuning-free plug-in methods~\cite{si2023freeu, zhao2023magicfusion, cao2023masactrl} and even a learnable method~\cite{hao2023vico}. As shown in Figure~\ref{qual:teaser}, DreamMatcher is effective even in extreme non-rigid personalization scenarios. We further validate the robustness of our method in challenging personalization scenarios. The ablation studies confirm our design choices and emphasize the effectiveness of each component.

\begin{figure*}[t]
    \begin{center}

    \includegraphics[width=0.99\textwidth]{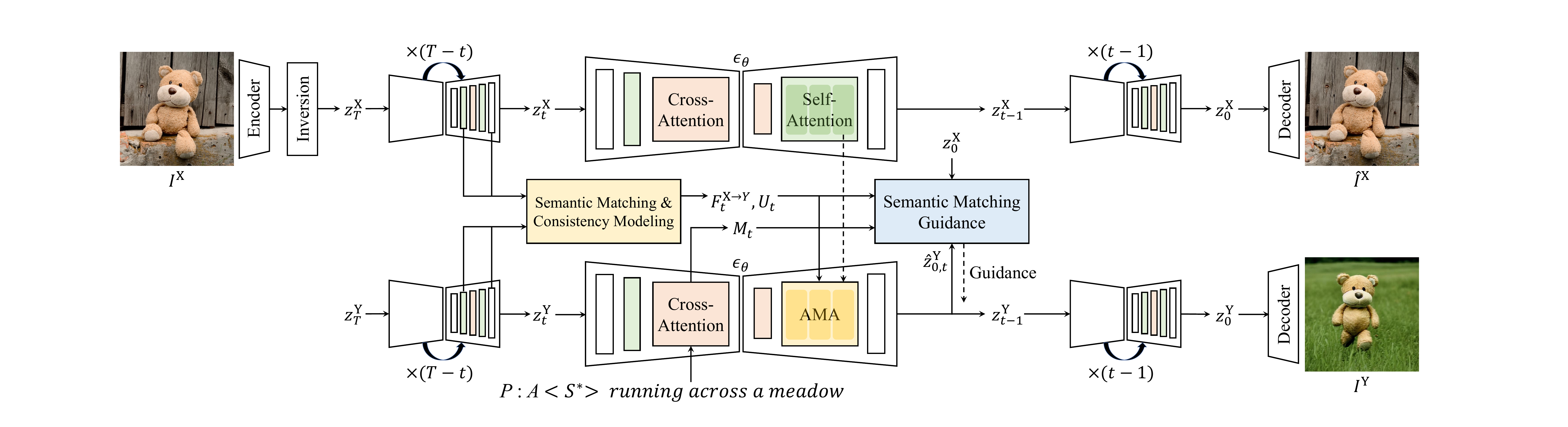} 
    \end{center}
    \vspace{-15pt}
    \caption{\textbf{Overall architecture:} Given a reference image $I^{X}$, appearance matching self-attention (AMA) aligns the reference appearance into the fixed target structure in self-attention module of pre-trained personalized model $\epsilon_{\theta}$. This is achieved by explictly leveraging reliable semantic matching from reference to target. Furthermore, semantic matching guidance enhances the fine-grained details of the subject in the generated images.
}
    \label{ill:overview}
    \vspace{-10pt}
\end{figure*}

\section{Related Work}
\label{sec:related_works}
\vspace{-5pt}
\paragrapht{Optimization-based T2I Personalization.}
Given a handful of images, T2I personalization aims to generate new image variations of the given concept that are consistent with the target prompt. Earlier diffusion-based techniques~\cite{gal2022image, voynov2023p+, ruiz2023dreambooth, kumari2023multi, han2023svdiff, dong2022dreamartist} encapsulate the given concept within the textual domain, typically represented by a specific token. Textual Inversion~\cite{gal2022image} optimizes a textual embedding and synthesizes personalized images by integrating the token with the target prompt. DreamBooth~\cite{ruiz2023dreambooth} proposes optimizing all parameters of the denoising U-Net based on a specific token and the class category of the subject. Several works~\cite{kumari2023multi, han2023svdiff, hao2023vico, chen2023disenbooth, tewel2023key, ruiz2023hyperdreambooth, xiang2023closer, liu2023cones} focus on optimizing weight subsets or an additional adapter for efficient optimization and better conditioning. For example, CustomDiffusion~\cite{kumari2023multi} fine-tunes only the cross-attention layers in the U-Net, while ViCo~\cite{hao2023vico} optimizes an additional image encoder. Despite promising results, the aforementioned approaches often fail to accurately mimic the appearance of the subject. 

\vspace{3pt}
\paragrapht{Training-based T2I Personalization.}
Several studies~\cite{jia2023taming, shi2023instantbooth, chen2023subject, xiao2023fastcomposer, gal2023designing, wei2023elite, chen2023photoverse, su2023identity, li2023blip} have shifted their focus toward training a T2I personalized model with large text-image pairs. For instance, Taming Encoder~\cite{jia2023taming}, InstantBooth~\cite{shi2023instantbooth}, and FastComposer~\cite{xiao2023fastcomposer} train an image encoder, while SuTI~\cite{chen2023subject} trains a separate network. While these approaches circumvent fine-tuning issues, they necessitate extensive pre-training with a large-scale dataset.

\vspace{3pt}
\paragrapht{Plug-in Subject-driven T2I Synthesis.}
Recent studies~\cite{cao2023masactrl, si2023freeu, zhao2023magicfusion, mou2023dragondiffusion, gu2023photoswap, seo2023midms} aim to achieve subject-driven T2I personalization or non-rigid editing without the need for additional fine-tuning or training. Specifically, MasaCtrl~\cite{cao2023masactrl} leverages dual-branch pre-trained diffusion models to incorporate image features from the reference branch into the target branch. FreeU~\cite{si2023freeu} proposes reweighting intermediate feature maps from a pre-trained personalized model~\cite{ruiz2023dreambooth}, based on frequency analysis. MagicFusion~\cite{zhao2023magicfusion} introduces a noise blending method between a pre-trained diffusion model and a T2I personalized model~\cite{ruiz2023dreambooth}. DreamMatcher is in alignment with these methods, designed to be compatible with any off-the-shelf T2I personalized models, thereby eliminating additional fine-tuning or training.

\section{Preliminary}
\vspace{-5pt}
\label{sec:preliminary}
\subsection{Latent Diffusion Models}
\vspace{-5pt}
Diffusion models~\cite{ho2020denoising, song2020denoising} generate desired data samples from Gaussian noise through a gradual denoising process. Latent diffusion models~\cite{rombach2022high} perform this process in the latent space projected by an autoencoder, instead of RGB space. Specifically, an encoder maps an RGB image $x_0$ into a latent variable $z_0$ and a decoder then reconstructs it back to $x_0$. In forward diffusion process, Gaussian noise is gradually added to the latent $z_t$ at each time step $t$ to produce the noisy latent $z_{t+1}$. In reverse diffusion process, the neural network $\epsilon_{\theta}(z_t, t)$ denoises $z_{t}$ to produce $z_{t-1}$ with the time step $t$. By iteratively sampling $z_{t-1}$, Gaussian noise $z_T$ is transformed into latent $z_0$. The denoised $z_0$ is converted back to $x_0$ using the decoder. When the condition, e.g., text prompt $P$, is added, $\epsilon_{\theta}(z_t, t, P)$ generates latents that are aligned with the text descriptions.

\subsection{Self-Attention in Diffusion Models}
\label{sec:self-attention}
\vspace{-5pt}
Diffusion model is often based on a U-Net architecture that includes residual blocks, cross-attention modules, and self-attention modules~\cite{ho2020denoising, song2020denoising,rombach2022high}. The residual block processes the features from preceding layers, the cross-attention module integrates these features with the condition, e.g., text prompt, and the self-attention module aggregates image features themselves through the attention operation.

Specifically, the self-attention module projects the image feature at time step $t$ into queries $Q_t$, keys $K_t$, and values $V_t$. The resulting output from this module is defined by:
\begin{equation}
\label{equ:attention}
\mathrm{SA}(Q_t, K_t, V_t) = \mathrm{Softmax}\left(\frac{{Q_t}{K^T_t}}{\sqrt{d}}\right)V_t.
\end{equation}
Here, $\mathrm{Softmax}(\cdot)$ is applied over the keys for each query. $ Q_t \in \mathbb{R}^{h \times w \times d}$, $K_t \in \mathbb{R}^{h \times w \times d}$, and $V_t \in \mathbb{R}^{h \times w \times d}$ are the projected matrices, where $h$, $w$, and $d$ refer to the height, width, and channel dimensions, respectively. As analyzed in~\cite{tumanyan2023plug, tewel2023key, balaji2022ediffi, hertz2022prompt}, we view the self-attention module as two distinct paths: the \textit{structure} and \textit{appearance} paths. More specifically, the structure path is defined by the similarities $\mathrm{Softmax}({Q_tK_t^T}/{\sqrt{d}})$, which controls the spatial arrangement of image elements. The values $V_t$ constitute the appearance path, injecting visual attributes such as colors, textures, and shapes, to each corresponding element within the image.

\section{Method}
\vspace{-5pt}
\label{sec:method}
 \begin{figure}[t]
    \centering
    \includegraphics[width=0.45\textwidth]{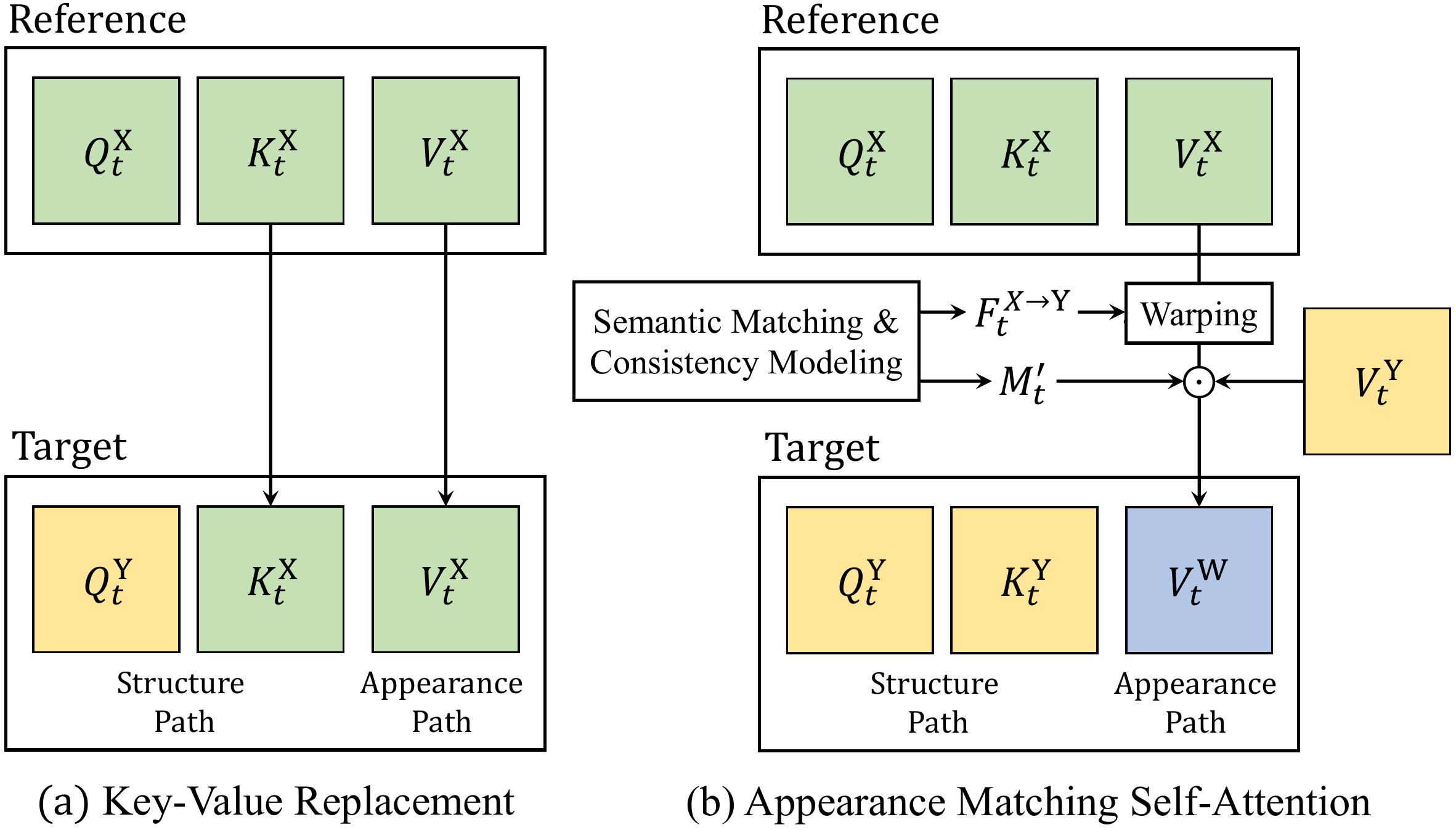}
    \vspace{-5pt}
    \caption{\textbf{Comparison between (a) key-value replacement~\cite{cao2023masactrl, mou2023dragondiffusion, chen2023anydoor, chen2023fec, huang2023kv, khandelwal2023infusion} and (b) appearance matching self-attention (AMA):} AMA aligns the reference appearance path toward the fixed target structure path through explicit semantic matching and consistency modeling.}
    \vspace{-10pt}
\label{ill:conceptual}
\end{figure}

Given a set of $n$ reference images $\mathcal{X}=\{I^X_n\}_{n1}^{N}$ , conventional methods~\cite{gal2022image, ruiz2023dreambooth, kumari2023multi} personalize the T2I models $\epsilon_{\theta}(\cdot)$ with a specific text prompt for the subject (e.g., $\langle S^{*} \rangle$). In inference, $\epsilon_{\theta}(\cdot)$ can generate novel scenes from random noises through iterative denoising processes with the subject aligned by the target prompt (e.g., \textit{A $\langle S^{*} \rangle$ in the jungle}). However, they often fail to accurately mimic the subject appearance because text embeddings lack the spatial expressivity to represent the visual attributes of the subject~\cite{hao2023vico, radford2021learning}.  In this paper, with a set of reference images $\mathcal{X}$ and a target text prompt $P$, we aim to enhance the subject appearance in the personalized image $I^Y$, while preserving the detailed target structure directed by the prompt $P$. DreamMatcher comprises a reference-target dual-branch framework. $I^{X}$ is inverted to $z^{X}_{T}$ via DDIM inversion~\cite{song2020denoising} and then reconstructed to $\hat{I}^{X}$, while $I^{Y}$ is generated from a random Gaussian noise $z^{Y}_{T}$ guided by $P$. At each time step, the self-attention module from the reference branch projects image features into queries $Q^{X}_{t}$, $K^{X}_{t}$, and $V^{X}_{t}$, while the target branch produces $Q^{Y}_{t}$, $K^{Y}_{t}$, and $V^{Y}_{t}$. The reference appearance $V^{X}_{t}$ is then transferred to the target denoising U-Net through its self-attention module. The overall architecture of DreamMatcher is illustrated in Figure~\ref{ill:overview}.

\subsection{Appearance Matching Self-Attention}
\label{sec:AMA} 
\vspace{-5pt}
As illustrated in Figure~\ref{ill:conceptual}, we propose an appearance matching self-attention (AMA) which manipulates only the appearance path while retaining the pre-trained target structure path, in order to enhance subject expressivity while preserving the target prompt-directed layout. 

However, naively swapping the target appearance $V^{Y}_{t}$ with that from the reference $V^{X}_{t}$, which reformulates Equation~\ref{equ:attention}, results in structure-appearance misalignment:
\begin{equation}
\label{equ:sama_attention}
\mathrm{SA}(Q^{Y}_t, K^{Y}_t, V^{X}_t) = \mathrm{Softmax}\left(\frac{{Q^{Y}_t}{(K^{Y}_t)^T}}{\sqrt{d}}\right)V^{X}_t.
\end{equation}
To solve this, we propose a matching-aware value injection method that leverages semantic matching to accurately align the reference appearance $V^{X}_t$ with the fixed target structure $\mathrm{Softmax}({{Q^{Y}_t}{(K^{Y}_t)^T}}/{\sqrt{d}})$. Specifically, AMA warps the reference values $V^{X}_t$ by the estimated semantic correspondence $F^{X \rightarrow Y}_{t}$ from reference to target, which is a dense displacement field~\cite{truong2020glu, truong2020gocor, truong2023pdc, cho2021cats, nam2023diffmatch} between semantically identical locations in both images. The warped reference values $V^{X \rightarrow Y}_{t}$ are formulated by:
\begin{equation}
\label{eq:warping}
    V^{X \rightarrow Y}_{t} = \mathcal{W}(V^{X}_{t};{{F}^{X \rightarrow Y}_{t}}),
\end{equation}
where $\mathcal{W}$ represents the warping operation~\cite{truong2021warp}.

In addition, it is crucial to isolate only the matched reference appearance and filter out outliers. This is because typical personalization scenarios often involve occlusions, different viewpoints, or background changes that are not present in the reference images, as shown in Figure~\ref{qual:teaser}. To achieve this, previous methods~\cite{cao2023masactrl, hao2023vico} use a foreground mask $M_t$ to focus only on the subject foreground and handle background variations. $M_t$ is obtained from the averaged cross-attention map for the subject text prompt (e.g., $\langle S^{*} \rangle$). With these considerations, Equation~\ref{eq:warping} can be reformulated as follows:
\begin{equation}
\label{eq:warping_with_uc}
    V^{W}_{t} = V^{X \rightarrow Y}_{t} \odot M_t + V^{Y}_{t} \odot (1-M_t),
\end{equation}
where $\odot$ represents Hadamard product~\cite{horn1990hadamard}. 

AMA then implants $V^{W}_{t}$ into the fixed target structure path through the self-attention module. Equation~\ref{equ:sama_attention} is reformulated as:
\begin{equation}
\label{equ:new_sama_attention}
\mathrm{AMA}(Q^{Y}_t, K^{Y}_t, V^{W}_t) = \mathrm{Softmax}\left(\frac{{Q^{Y}_t}{(K^{Y}_t)^T}}{\sqrt{d}}\right)V^{W}_t.
\end{equation}

\begin{figure}[t]
    \centering
    \includegraphics[width=0.45\textwidth]{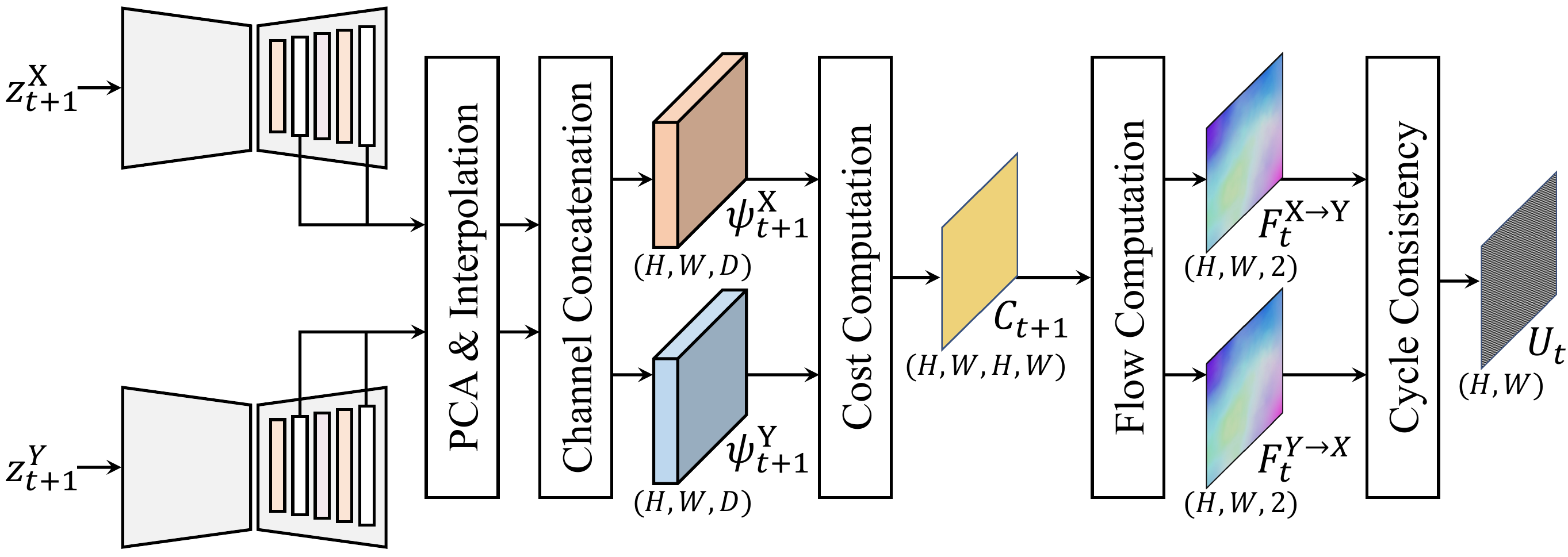}
    \vspace{-5pt}
    \caption{\textbf{Semantic matching and consistency modeling:} We leverage internal diffusion features at each time step to find semantic matching $F_{t}^{\mathrm{X\rightarrow Y}}$ between reference and target. Additionally, we compute the confidence map of the predicted matches $U_{t}$ through cycle-consistency.}
    \vspace{-10pt}
\label{ill:matching_module}
\end{figure}

In our framework, we find semantic correspondence between reference and target, aligning with standard semantic matching workflows~\cite{truong2020glu, truong2020gocor, cho2021cats, cho2022cats++, hong2022neural, nam2023diffmatch}. Figure~\ref{ill:matching_module} provides a detailed schematic of the proposed matching process. In the following, we will explain the process in detail.

\vspace{3pt}
\paragrapht{Feature Extraction.} 
Classical matching pipelines~\cite{truong2020glu, truong2020gocor, cho2021cats, cho2022cats++, hong2022neural, nam2023diffmatch} contain pre-trained feature extractors~\cite{chatfield2014return, he2016deep, oquab2023dinov2} to obtain feature descriptors $\psi^{X}$ and $\psi^{Y}$ from image pairs $I^{X}$ and $I^{Y}$. However, finding good features tailored for T2I personalization is not trivial due to the noisy nature of estimated target images in reverse diffusion process, requiring additional fine-tuning of the existing feature extractors.
To address this, we focus on the diffusion feature space~\cite{tang2023emergent, zhang2023tale} in the pre-trained T2I model itself to find a semantic matching tailored for T2I personalization.

Let $\epsilon_{\theta,l}(\cdot,t+1)$ denote the output of the $l$-th decoder layer of the denoising U-Net~\cite{ho2020denoising} $\epsilon_{\theta}$ at time step $t+1$. Given the latent $z_{t+1}$ with time step $t+1$ and text prompt $P$ as inputs, we extract the feature descriptor $\psi_{t+1,l}$ from the $l$-th layer of the U-Net decoder. The process is given by:
\begin{equation} 
\psi_{t+1,l} = \epsilon_{\theta, l}(z_{t+1},\, t+1,\, P),
\end{equation}
where we obtain $\psi_{t+1,l}^{X}$ and $\psi_{t+1,l}^{Y}$ from $z^{X}_{t+1}$ and $z^{Y}_{t+1}$, respectively. For brevity, we will omit $l$ in the following discussion. 

\begin{figure}[t]
    \centering
    \includegraphics[width=0.45\textwidth]{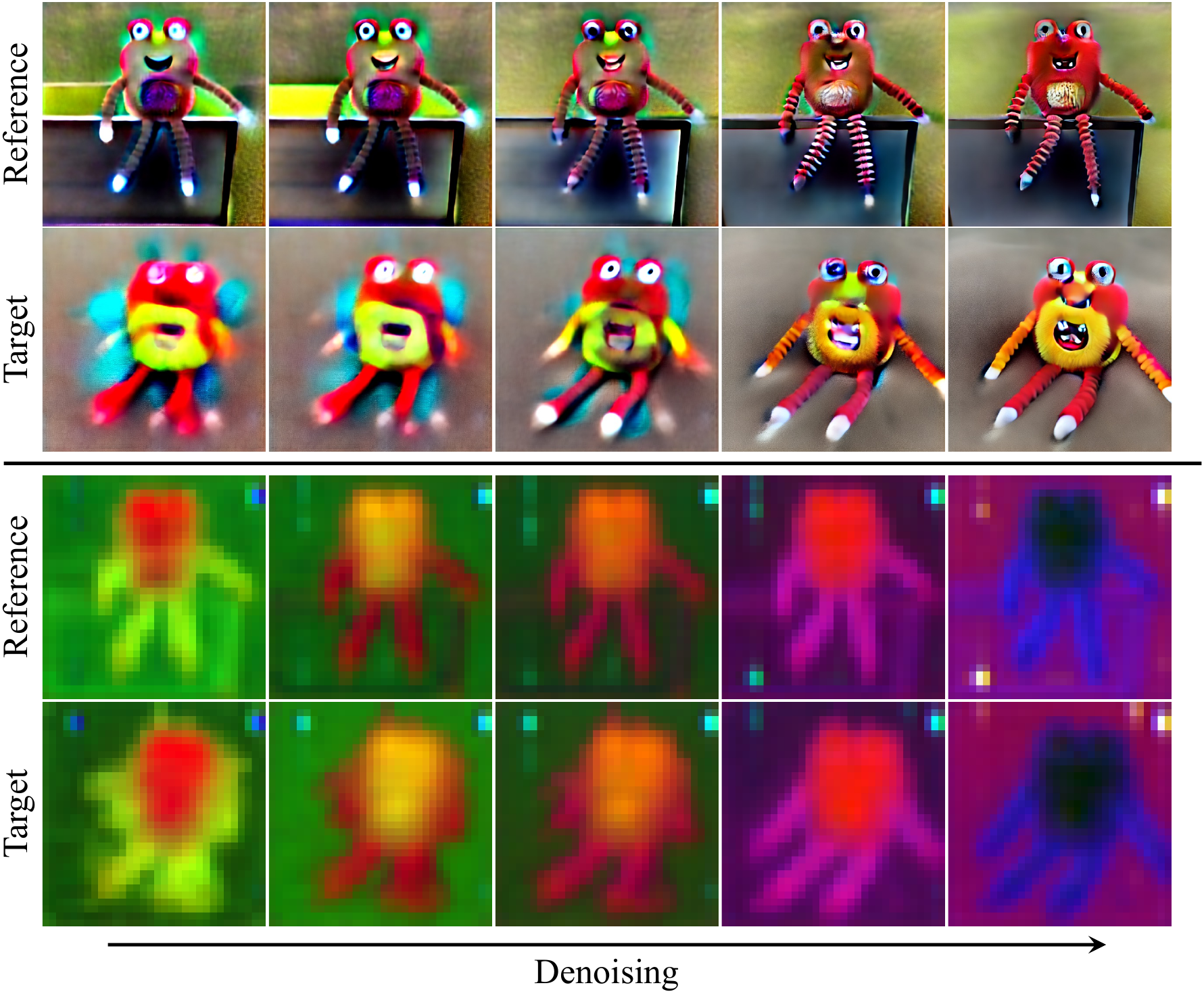}
    \vspace{-5pt}    \caption{\textbf{Diffusion feature visualization:} Upper displays intermediate estimated images of reference and target, with the target generated by DreamBooth~\cite{ruiz2023dreambooth} using the prompt \textit{A $\langle S^* \rangle$ on the beach}. Lower visualizes the three principal components of intermediate diffusion features. The similar semantics share similar colors.}
    \vspace{-10pt}
    \label{qual:feat_vis}
\end{figure}

To explore the semantic relationship within the diffusion feature space between reference and target, 
Figure~\ref{qual:feat_vis} visualizes the relation between $\psi_{t+1}^{X}$ and $\psi_{t+1}^{Y}$ at different time steps using principal component analysis (PCA)~\cite{pearson1901liii}. We observe that the foreground subjects share similar semantics, even they have different appearances, as the target image from the pre-trained personalized model often lacks subject expressivity. This observation inspires us to leverage the internal diffusion features to establish semantic matching between estimated reference and target at each time step of sampling phase.

Based on this, we derive $ \psi_{t+1} \in \mathbb{R}^{H \times W \times D} $ by combining PCA features from different layers using channel concatenation, where $D$ is the concatenated channel dimension. Detailed analysis and implementation on feature extraction is provided in Appendix ~\ref{supp:ama}. 

\vspace{3pt}
\paragrapht{Flow Computation.} Following conventional methods~\cite{truong2020glu, truong2020gocor, truong2023pdc, cho2021cats, hong2022neural}, we build the matching cost by calculating the pairwise cosine similarity between feature descriptors for both the reference and target images. For given $\psi^{X}_{t+1}$ and $\psi^{Y}_{t+1}$ at time step $t+1$, the matching cost $C_{t+1}$ is computed by taking dot products between all positions in the feature descriptors. This is formulated as: 
\begin{equation}\label{equ:cost} 
C_{t+1}(i, j) = \frac{{\psi^{{X}}_{t+1}}(i) \cdot \psi^{{Y}}_{t+1}(j)}{\|\psi^{{X}}_{t+1}(i)\| \|{\psi^{{Y}}_{t+1}}(j)\|},
\end{equation}
where $i, j \in [0, H) \times [0, W)$, and $\|\cdot\|$ denotes $l$-2 normalization.

Subsequently, we derive the dense displacement field from the reference to the target at time step $t$, denoted as $F^{{X \rightarrow Y}}_{t} \in \mathbb{R}^{H \times W\times 2}$, using the argmax operation~\cite{cho2021cats} on the matching cost $C_{t+1}$. Figure~\ref{qual:matching_module}(c) shows the warped reference image obtained using the predicted correspondence $F^{{X \rightarrow Y}}_{t}$ between $\psi^{X}_{t+1}$ and $\psi^{Y}_{t+1}$ in the middle of the generation process. This demonstrates that the correspondence is established reliably in reverse diffusion process, even in intricate non-rigid target contexts that include large displacements, occlusions, and novel-view synthesis.

\subsection{Consistency Modeling}
\begin{figure}[t]
    \centering
    \includegraphics[width=0.45\textwidth]{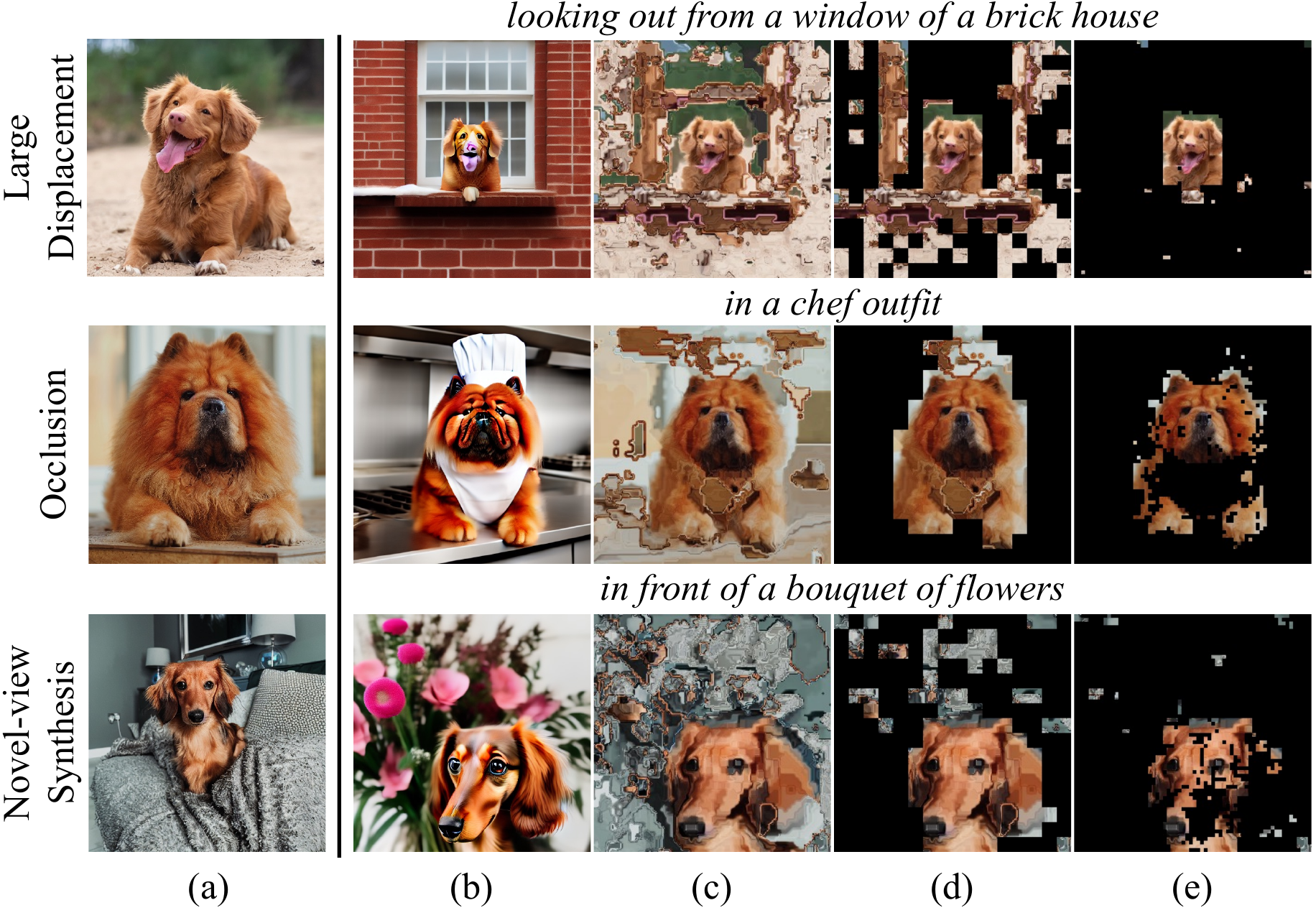}
    \vspace{-5pt}
    \caption{\textbf{Correspondence visualization:} (a) Reference image. (b) Estimated target image from DreamBooth~\cite{ruiz2023dreambooth} at 50\% of the reverse diffusion process. (c) Warped reference image based on predicted correspondence $F^{X \rightarrow Y}_t$. (d) Warped reference image combined with foreground mask $M_t$. (e) Warped reference image combined with both $M_t$ and confidence mask $U_t$.}
    \vspace{-10pt}
\label{qual:matching_module}
\end{figure}

\vspace{-5pt}
As depicted in Figure~\ref{qual:matching_module}(d), the forground mask $M_t$ is insufficient to address occlusions and background clutters, (e.g., \textit{a chef outfit} or \textit{a bouquet of flowers}), as these are challenging to distinguish within the cross-attention module.

To compensate for this, we introduce a confidence mask $U_t$ to discard erroneous correspondences, thus preserving detailed target structure. Specifically, we enforce a cycle consistency constraint~\cite{jiang2021cotr}, simply rejecting any correspondence greater than the threshold we set. In other words, we only accept correspondences where a target location $x$ remains consistent when a matched reference location, obtained by ${F}^{Y \rightarrow X}_t$, is re-warped using ${F}^{X \rightarrow Y}_t$. We empirically set the threshold proportional to the target foreground area. This is formulated by:
\begin{equation}
U_t(x) = 
\begin{cases} 
1, & \text{if } \left\| \mathcal{W} \left( F^{Y \rightarrow X}_t ; F^{X \rightarrow Y}_t \right)(x) \right\| < \gamma \lambda_c, \\
0, & \text{otherwise},
\end{cases}
\end{equation}
where $\|\cdot\|$ denotes a $l$-2 norm, and $\mathcal{W}$ represents the warping operation \cite{truong2021warp}. ${F}^{Y \rightarrow X}_t$ indicates the reverse flow field of its forward counterpart, ${F}^{X \rightarrow Y}_t$. $\gamma$ is a scaling factor designed to be proportional to foreground area, and $\lambda_{c}$ is a hyperparameter. More details are available in Appendix~\ref{supp:cf}.

Finally, we define a semantic-consistent mask $M'_t$ by combining $M_t$ and $U_t$ using the Hadamard product~\cite{horn1990hadamard}, so that $M_t$ coarsely captures the foreground subject, while $U_t$ finely filters out unreliable matches and preserves the fine-grained target context. As shown in Figure~\ref{qual:matching_module}(e), our network selectively incorporates only the confident matches, effectively addressing intricate non-rigid scenarios. 

We now apply a confidence-aware modification to appearance matching self-attention in Equation~\ref{eq:warping_with_uc}, by replacing $M_t$ with $M'_t$.

\subsection{Semantic Matching Guidance}
\label{sec:matching_guidance}
\vspace{-5pt}
Our method uses intermediate reference values $V^{X}_t$ at each time step. However, we observe that in early time steps, these noisy values may lack fine-grained subject details, resulting in suboptimal results. To overcome this, we further introduce a sampling guidance technique, named semantic matching guidance, to provide rich reference semantics in the middle of the target denoising process. 

In terms of the score-based generative models~\cite{song2019generative, song2020score}, the guidance function $g$ steers the target images towards higher likelihoods. The updated direction $\hat{\epsilon}_t$ at time step $t$ is defined as follows~\cite{epstein2023diffusion}:
\begin{equation}
\hat{\epsilon}_t = \epsilon_{\theta}(z_t, t, P) - \lambda_{g} \sigma_t \nabla_{{z}_t}g(z_t,t, P),
\end{equation}
where $\lambda_{g}$ is a hyperparameter that modulates the guidance strength, and $\sigma_t$ represents the noise schedule parameter at time step $t$.

We design the guidance function $g$ using $z^{X}_0$ from DDIM inversion~\cite{song2020denoising}, which encapsulates detailed subject representation at the final reverse step $t=0$. At each time step $t$, $z^{X}_{0}$ is transformed to align with the target structure through $F^{X \rightarrow Y}_{t}$, as follows:
\begin{equation}
\label{equ:guidance_warping}
    z^{X \rightarrow Y}_{0, t} = \mathcal{W}(z^{X}_{0};{{F}^{X \rightarrow Y}_{t}}).
\end{equation}
The guidance function $g_t$ at time step $t$ is then defined as the pixel-wise difference between the aligned $ z^{X \rightarrow Y}_{0, t}$ and the target latent $\hat{z}^{Y}_{0, t}$ which is calculated by reparametrization trick~\cite{song2020denoising}, taking into account the semantic-consistent mask $M'_t$ :
\begin{equation}
    g_t = \frac{1}{|M'_t|} \sum_{i \in M'_t}\left\| z^{X \rightarrow Y}_{0,t}(i) - \hat{z}^{Y}_{0,t}(i)\right\|,
\end{equation}
where $\|\cdot\|$ denotes a $l$-2 norm.

Note that our approach differs from existing methods~\cite{epstein2023diffusion, mou2023dragondiffusion, bansal2023universal} that provide coarse appearance guidance by calculating the average feature difference between foregrounds. Instead, we leverage confidence-aware semantic correspondence to offer more precise and pixel-wise control.

\section{Experiments}
\vspace{-5pt}
\begin{figure*}[!ht]
    \begin{center}
        \includegraphics[width=0.95\textwidth]{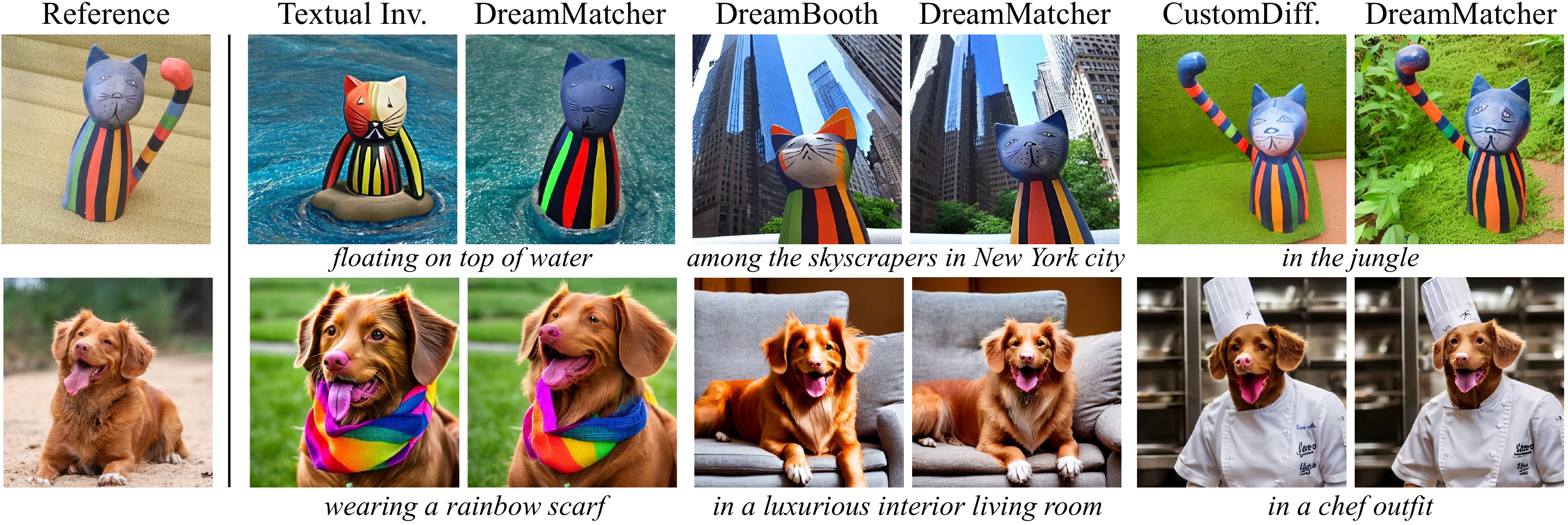} 
    \end{center}
    \vspace{-20pt}
\caption{\textbf{Qualitative comparison with baselines:} We compare DreamMatcher with three different baselines, Textual Inversion~\cite{gal2022image}, DreamBooth~\cite{ruiz2023dreambooth}, and CustomDiffusion~\cite{kumari2023multi}.}
    \label{qual:base}
     \vspace{-5pt}
\end{figure*}

\label{sec:experiments}
\subsection{Experimental Settings}
\vspace{-5pt}
\paragrapht{Dataset.} ViCo~\cite{hao2023vico} gathered an image-prompt dataset from previous works~\cite{gal2022image, ruiz2023dreambooth, kumari2023multi}, comprising 16 concepts and 31 prompts. We adhered to the ViCo dataset and evaluation settings, testing 8 samples per concept and prompt, for a total of 3,969 images. To further evaluate the robustness of our method in complex non-rigid personalization scenarios, we created a prompt dataset divided into three categories: large displacements, occlusions, and novel-view synthesis. This dataset includes 10 prompts for large displacements and occlusions, and 4 for novel-view synthesis, all created using ChatGPT~\cite{openai2023gpt4}. The detailed procedure and the prompt list are in the Appendix~\ref{supp:dataset}.

\vspace{3pt}
\paragrapht{Baseline and Comparison.} DreamMatcher is designed to be compatible with any T2I personalized models. We implemented our method using three baselines: Textual Inversion~\cite{gal2022image}, DreamBooth~\cite{ruiz2023dreambooth}, and CustomDiffusion~\cite{kumari2023multi}. We benchmarked DreamMatcher against previous tuning-free plug-in models, FreeU~\cite{si2023freeu} and MagicFusion~\cite{zhao2023magicfusion}, and also against the optimization-based model, ViCo~\cite{hao2023vico}. Note that additional experiments, including DreamMatcher on Stable Diffusion or DreamMatcher for multiple subject personalization, are provided in Appendix~\ref{supp:analysis}.

\vspace{3pt}
\paragrapht{Evaluation Metric.}
Following previous studies~\cite{gal2022image, ruiz2023dreambooth, kumari2023multi, hao2023vico}, we evaluated subject and prompt fidelity. For subject fidelity, we adopted the CLIP~\cite{radford2021learning} and DINO~\cite{caron2021emerging} image similarity, denoted as $I_\mathrm{CLIP}$ and $I_\mathrm{DINO}$, respectively. For prompt fidelity, we adopted the CLIP image-text similarity $T_\text{CLIP}$, comparing visual features of generated images to textual features of their prompts, excluding placeholders. Further details on evaluation metrics are in the Appendix~\ref{supp:metrics}.

\vspace{3pt}
\paragrapht{User Study.} We conducted a user study comparing DreamMatcher to previous works~\cite{hao2023vico, si2023freeu, zhao2023magicfusion}. Participants evaluated the generated images from different methods based on subject and prompt fidelity. 45 users responded to 32 comparative questions, totaling 1440 responses. Samples were chosen randomly from a large, unbiased pool. Additional details on the user study are in Appendix~\ref{supp:user_study}.
\begin{table}[t]
\centering
\resizebox{0.7\columnwidth}{!}{%
\begin{tabular}{l|ccc}
    \toprule
     Method & $I_\mathrm{DINO}$ $\uparrow$ & $I_\mathrm{CLIP}$ $\uparrow$  & $T_\mathrm{CLIP}$ $\uparrow$ \\
    \midrule
    Textual Inversion~\cite{gal2022image}  &0.529 &0.762 & \textbf{0.220}\\
    DreamMatcher &\textbf{0.588} \textcolor{blue}{(+11.2\%)} &\textbf{0.778} \textcolor{blue}{(+2.1\%)}  & 0.217 \textcolor{red}{(-1.4\%)}\\
    \midrule
    DreamBooth~\cite{ruiz2023dreambooth}   &0.638 &0.808 & \textbf{0.237}\\
    DreamMatcher   &\textbf{0.680} \textcolor{blue}{(+6.6\%)} &\textbf{0.821} \textcolor{blue}{(+1.6\%)}  & 0.231 \textcolor{red}{(-2.5\%)} \\
    \midrule
    CustomDiffusion~\cite{kumari2023multi} &0.667 &0.810  & 0.218  \\
    DreamMatcher&\textbf{0.700} \textcolor{blue}{(+4.9\%)}&\textbf{0.821} \textcolor{blue}{(+1.4\%)}  & \textbf{0.223} \textcolor{blue}{(+2.3\%)} \\
    \bottomrule
\end{tabular}%
}
\vspace{-5pt}
\caption{\textbf{Quantitative comparison with different baselines.}}
\vspace{-15pt}
\label{quan:base}
\end{table}

\subsection{Results}
\vspace{-5pt}
\paragrapht{Comparison with Baselines.} 
Table~\ref{quan:base} and Figure~\ref{qual:base} summarize the quantitative and qualitative comparisons with different baselines. The baselines~\cite{gal2022image, ruiz2023dreambooth, kumari2023multi} often lose key visual attributes of the subject such as colors, texture, or shape due to the limited expressivity of text embeddings. In contrast, DreamMatcher significantly outperforms these baselines by a large margin in subject fidelity $I_\mathrm{DINO}$ and $I_\mathrm{CLIP}$, while effectively preserving prompt fidelity $T_\mathrm{CLIP}$. As noted in~\cite{hao2023vico, ruiz2023dreambooth}, we want to highlight that $I_\mathrm{DINO}$ better reflects subject expressivity, as it is trained in a self-supervised fashion, thus distinguishing the difference among objects in the same category. Additionally, we wish to note that better prompt fidelity does not always reflect in $T_\mathrm{CLIP}$. $T_\mathrm{CLIP}$ is reported to imperfectly capture text-image alignment and has been replaced by the VQA-based evaluation~\cite{ghosh2023geneval, yarom2023you}, implying its slight performance drop is negligible. More results are provided in Appendix~\ref{supp:comp_base}.

\begin{table}[t]
\centering
\resizebox{0.6\columnwidth}{!}{%
\begin{tabular}{l|ccc}
    \toprule
     Method & $I_\mathrm{DINO}$ $\uparrow$ & $I_\mathrm{CLIP}$ $\uparrow$ &  $T_\mathrm{CLIP}$ $\uparrow$\\ 
    \midrule
    MagicFusion~\cite{zhao2023magicfusion} &  0.632 & 0.811 & 0.233 \\
    FreeU~\cite{si2023freeu} & 0.632 & 0.806 & \textbf{0.236} \\ \midrule
    DreamMatcher & \textbf{0.680} &\textbf{0.821} & 0.231 \\ 

    \bottomrule
\end{tabular}%
}
\vspace{-5pt}
\caption{\textbf{Quantitative comparison with tuning-free methods.} For this comparison, we used DreamBooth~\cite{ruiz2023dreambooth} as our baseline.}
\vspace{-5pt}
\label{quan:sota}
\end{table}

\begin{table}[t]
\centering
\resizebox{0.6\columnwidth}{!}{%
\begin{tabular}{l|ccc}
    \toprule
     Method & $I_\mathrm{DINO}$ $\uparrow$ & $I_\mathrm{CLIP}$ $\uparrow$  & $T_\mathrm{CLIP}$ $\uparrow$\\
    \midrule
    
    MagicFusion~\cite{zhao2023magicfusion} & 0.622 & 0.814 & 0.235\\
    FreeU~\cite{si2023freeu} &0.611 & 0.803 & \textbf{0.242} \\
    
    \midrule
    
    DreamMatcher  &\textbf{0.655} &\textbf{0.818} &0.239 \\

    \bottomrule
\end{tabular}%
}
\vspace{-5pt}
\caption{\textbf{Quantitative comparison in challenging dataset.} For this comparison, we used DreamBooth~\cite{ruiz2023dreambooth} as our baseline.}
\vspace{-5pt}
\label{quan:challenging}
\end{table}

\begin{table}[t]
\centering
\resizebox{0.6\columnwidth}{!}{%
\begin{tabular}{l|ccc}
    \toprule
     Method  & $I_\mathrm{DINO}$ $\uparrow$ & $I_\mathrm{CLIP}$ $\uparrow$ & $T_\mathrm{CLIP}$ $\uparrow$\\
    \midrule
    ViCo~\cite{hao2023vico}  &0.643& 0.816&\textbf{0.228} \\
    \midrule
   DreamMatcher &\textbf{0.700}&\textbf{0.821}&0.223\\

    \bottomrule
\end{tabular}%
}
\vspace{-5pt}
\caption{\textbf{Comparison with optimization-based method.} For this comparison, we used CustomDiffusion~\cite{kumari2023multi} as our baseline.}
\vspace{-15pt}
\label{quan:vico}
\end{table}

\begin{figure*}[t]
    \begin{center}
        \includegraphics[width=0.95\textwidth]{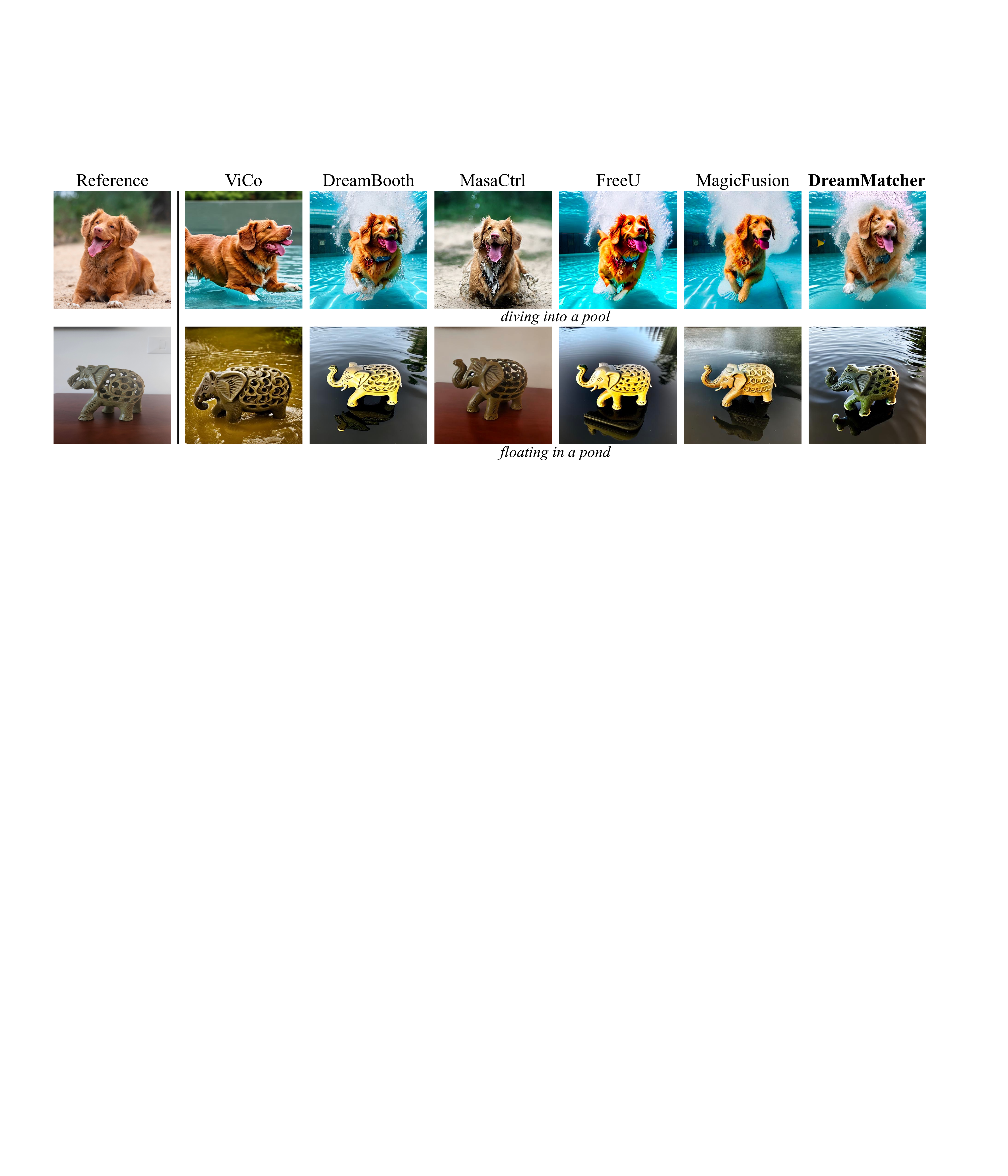} 
    \end{center}
    \vspace{-20pt}
    \caption{\textbf{Qualitative comparison with previous works}~\cite{hao2023vico, ruiz2023dreambooth, cao2023masactrl,
    si2023freeu, zhao2023magicfusion}\textbf{:} For this comparison, DreamBooth~\cite{ruiz2023dreambooth} was used as the baseline of MasaCtrl, FreeU, MagicFusion, and DreamMatcher.
}
    \vspace{-5pt}
    \label{qual:sota}
\end{figure*}

\begin{figure}[t]
    \centering
    \includegraphics[width=0.48\textwidth]{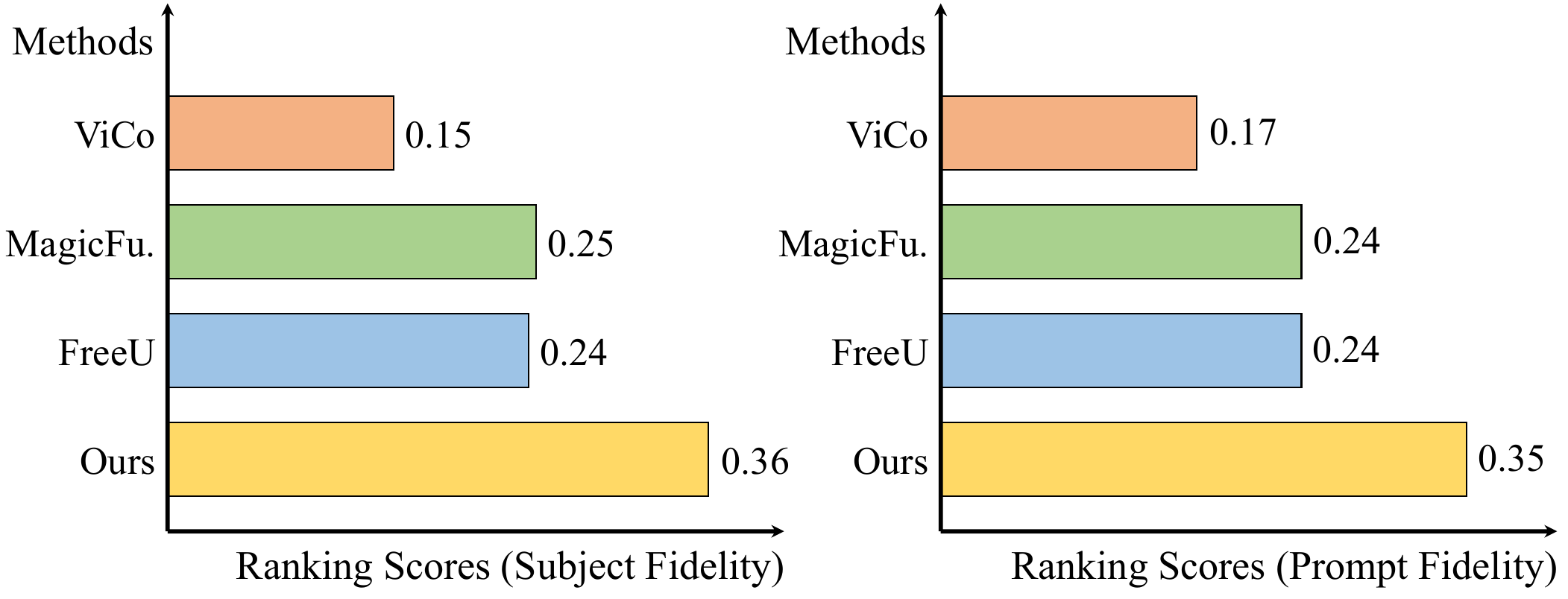}
    \vspace{-15pt}
    \caption{\textbf{User study.}}
    \vspace{-15pt}
    \label{qual:user_study}
\end{figure}

\vspace{3pt}
\paragrapht{Comparison with Plug-in Models.} 
We compared DreamMatcher against previous tuning-free plug-in methods, FreeU~\cite{si2023freeu} and MagicFusion~\cite{zhao2023magicfusion}. Both methods demonstrated their effectiveness when plugged into DreamBooth~\cite{ruiz2023dreambooth}. For a fair comparison, we evaluated DreamMatcher using DreamBooth as a baseline. As shown in Table~\ref{quan:sota} and Figure~\ref{qual:sota}, DreamMatcher notably outperforms these methods in subject fidelity, maintaining comparable prompt fidelity. The effectiveness of our method is also evident in Table~\ref{quan:challenging}, displaying quantitative results in challenging non-rigid personalization scenarios. This highlights the importance of semantic matching for robust performance in complex real-world personalization applications.

\vspace{3pt}
\paragrapht{Comparison with Optimization-based Models.}
We further evaluated DreamMatcher against the optimization-based model, ViCo~\cite{hao2023vico}, which fine-tunes an image adapter with 51.3M parameters. For a balanced comparison, we compared ViCo with DreamMatcher combined with CustomDiffusion~\cite{kumari2023multi}, configured with a similar count of trainable parameters (57.1M). Table~\ref{quan:vico} shows DreamMatcher notably surpasses ViCo in all subject fidelity metrics, without requiring extra fine-tuning. Figure~\ref{qual:sota} provides the qualitative comparison. More results are provided in Appendix~\ref{supp:comp_sota}.

\vspace{3pt}
\paragrapht{User Study.}
We also present the user study results in Figure~\ref{qual:user_study}, where DreamMatcher significantly surpasses all other methods in both subject and prompt fidelity. Further details are provided in Appendix~\ref{supp:user_study}.

\subsection{Ablation Study}
\begin{figure}[t]
    \centering
    \includegraphics[width=0.48\textwidth]{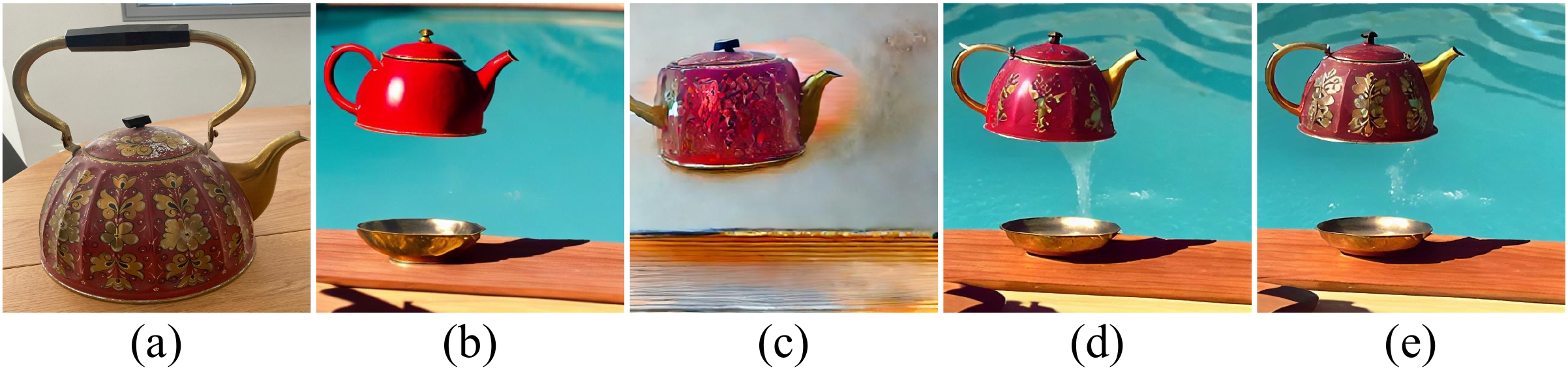}
    \vspace{-20pt}
    \caption{\textbf{Component analysis:} (a) reference image, (b) generated image by DreamBooth~\cite{ruiz2023dreambooth}, (c) with proposed semantic matching, (d) further combined with semantic-consistent mask, and (e) further combined with semantic matching guidance. }
    \vspace{-5pt}
    \label{qual:abl_comp}
\end{figure}

\vspace{-5pt}
In Figure~\ref{qual:abl_comp} and Table~\ref{quan:comp_analysis}, we demonstrate the effectiveness of each component in our framework. (b) and (I) present the results of the baseline, while (II) shows the results of key-value replacement, which fails to preserve the target structure and generates a static subject image. (c) and (III) display AMA using predicted correspondence, enhancing subject fidelity compared to (b) and (I), but drastically reducing prompt fidelity, as it could not filter out unreliable matches. This is addressed in (d) and (IV), which highlight the effectiveness of the semantic-consistent mask in significantly improving prompt fidelity, up to the baseline (I). Finally, the comparison between (d) and (e) demonstrate that semantic-matching guidance improves subject expressivity with minimal sacrifice in target structure, which is further evidenced by (V). More analyses, including a user study comparing DreamMatcher and MasaCtrl, are in Appendix~\ref{supp:analysis}.

    \begin{table}[t]
    \centering
    \resizebox{1
    \columnwidth}{!}{
    \begin{tabular}{c|l|ccc}
        \toprule
        & Component & $I_\mathrm{DINO}$ $\uparrow$ & $I_\mathrm{CLIP}$ $\uparrow$ &  $T_\mathrm{CLIP}$ $\uparrow$ \\
        \midrule
        (I) & Baseline (DreamBooth~\cite{ruiz2023dreambooth}) &0.638 &0.808 & 0.237 \\
        \midrule
        (II) & (I) + Key-Value Replacement (MasaCtrl~\cite{cao2023masactrl}) & 0.728 & 0.854 & 0.201 \\
        \midrule
        (III) & (I) + Semantic Matching &0.683&0.830&0.201 \\
        (IV) & (III) + Semantic-Consistent Mask (AMA)
          &0.676&0.818&0.232\\
        (V) & (IV) + Semantic Matching Guid. (Ours) &0.680 &0.821 & 0.231\\
        \bottomrule
    \end{tabular}
    }
    \vspace{-5pt}
    \caption{\textbf{Component analysis.} For this analysis, we used DreamBooth~\cite{ruiz2023dreambooth} for the baseline.}
    \vspace{-15pt}
    \label{quan:comp_analysis}
    \end{table}

\section{Conclusion}
\vspace{-5pt}
\label{sec:conclusion}
We present DreamMatcher, a tuning-free plug-in for text-to-image (T2I) personalization. DreamMatcher enhances appearance resemblance in personalized images by providing semantically aligned visual conditions, leveraging the generative capabilities of the self-attention module within pre-trained T2I personalized models. DreamMatcher pioneers the significance of semantically aligned visual conditioning in personalization, offering an effective solution within the attention framework. Experiments show that DreamMatcher enhances the personalization capabilities of existing T2I models, outperforming previous tuning-free plug-ins, even in complex scenarios.

\section{Acknowledgements} This research was supported by the MSIT, Korea (IITP-2024-2020-0-01819, ICT Creative Consilience Program, No.2021-0-02068, Artificial Intelligence Innovation Hub).

{
    \small
    \bibliographystyle{ieeenat_fullname}
    \bibliography{main}

\begin{thebibliography}{68}
\providecommand{\natexlab}[1]{#1}
\providecommand{\url}[1]{\texttt{#1}}
\expandafter\ifx\csname urlstyle\endcsname\relax
  \providecommand{\doi}[1]{doi: #1}\else
  \providecommand{\doi}{doi: \begingroup \urlstyle{rm}\Url}\fi

\bibitem[Balaji et~al.(2022)Balaji, Nah, Huang, Vahdat, Song, Kreis, Aittala, Aila, Laine, Catanzaro, et~al.]{balaji2022ediffi}
Yogesh Balaji, Seungjun Nah, Xun Huang, Arash Vahdat, Jiaming Song, Karsten Kreis, Miika Aittala, Timo Aila, Samuli Laine, Bryan Catanzaro, et~al.
\newblock ediffi: Text-to-image diffusion models with an ensemble of expert denoisers.
\newblock \emph{arXiv preprint arXiv:2211.01324}, 2022.

\bibitem[Bansal et~al.(2023)Bansal, Chu, Schwarzschild, Sengupta, Goldblum, Geiping, and Goldstein]{bansal2023universal}
Arpit Bansal, Hong-Min Chu, Avi Schwarzschild, Soumyadip Sengupta, Micah Goldblum, Jonas Geiping, and Tom Goldstein.
\newblock Universal guidance for diffusion models.
\newblock In \emph{Proceedings of the IEEE/CVF Conference on Computer Vision and Pattern Recognition}, pages 843--852, 2023.

\bibitem[Brooks et~al.(2023)Brooks, Holynski, and Efros]{brooks2023instructpix2pix}
Tim Brooks, Aleksander Holynski, and Alexei~A Efros.
\newblock Instructpix2pix: Learning to follow image editing instructions.
\newblock In \emph{Proceedings of the IEEE/CVF Conference on Computer Vision and Pattern Recognition}, pages 18392--18402, 2023.

\bibitem[Cao et~al.(2023)Cao, Wang, Qi, Shan, Qie, and Zheng]{cao2023masactrl}
Mingdeng Cao, Xintao Wang, Zhongang Qi, Ying Shan, Xiaohu Qie, and Yinqiang Zheng.
\newblock Masactrl: Tuning-free mutual self-attention control for consistent image synthesis and editing.
\newblock \emph{arXiv preprint arXiv:2304.08465}, 2023.

\bibitem[Caron et~al.(2021)Caron, Touvron, Misra, J{\'e}gou, Mairal, Bojanowski, and Joulin]{caron2021emerging}
Mathilde Caron, Hugo Touvron, Ishan Misra, Herv{\'e} J{\'e}gou, Julien Mairal, Piotr Bojanowski, and Armand Joulin.
\newblock Emerging properties in self-supervised vision transformers.
\newblock In \emph{Proceedings of the IEEE/CVF international conference on computer vision}, pages 9650--9660, 2021.

\bibitem[Chatfield et~al.(2014)Chatfield, Simonyan, Vedaldi, and Zisserman]{chatfield2014return}
Ken Chatfield, Karen Simonyan, Andrea Vedaldi, and Andrew Zisserman.
\newblock Return of the devil in the details: Delving deep into convolutional nets.
\newblock \emph{arXiv preprint arXiv:1405.3531}, 2014.

\bibitem[Chen et~al.(2023{\natexlab{a}})Chen, Zhang, Wang, Duan, Zhou, and Zhu]{chen2023disenbooth}
Hong Chen, Yipeng Zhang, Xin Wang, Xuguang Duan, Yuwei Zhou, and Wenwu Zhu.
\newblock Disenbooth: Disentangled parameter-efficient tuning for subject-driven text-to-image generation.
\newblock \emph{arXiv preprint arXiv:2305.03374}, 2023{\natexlab{a}}.

\bibitem[Chen et~al.(2023{\natexlab{b}})Chen, Zhao, Liu, Ding, Song, Wang, Wang, Yang, Liu, Du, et~al.]{chen2023photoverse}
Li Chen, Mengyi Zhao, Yiheng Liu, Mingxu Ding, Yangyang Song, Shizun Wang, Xu Wang, Hao Yang, Jing Liu, Kang Du, et~al.
\newblock Photoverse: Tuning-free image customization with text-to-image diffusion models.
\newblock \emph{arXiv preprint arXiv:2309.05793}, 2023{\natexlab{b}}.

\bibitem[Chen and Huang(2023)]{chen2023fec}
Songyan Chen and Jiancheng Huang.
\newblock Fec: Three finetuning-free methods to enhance consistency for real image editing.
\newblock \emph{arXiv preprint arXiv:2309.14934}, 2023.

\bibitem[Chen et~al.(2023{\natexlab{c}})Chen, Hu, Li, Rui, Jia, Chang, and Cohen]{chen2023subject}
Wenhu Chen, Hexiang Hu, Yandong Li, Nataniel Rui, Xuhui Jia, Ming-Wei Chang, and William~W Cohen.
\newblock Subject-driven text-to-image generation via apprenticeship learning.
\newblock \emph{arXiv preprint arXiv:2304.00186}, 2023{\natexlab{c}}.

\bibitem[Chen et~al.(2023{\natexlab{d}})Chen, Huang, Liu, Shen, Zhao, and Zhao]{chen2023anydoor}
Xi Chen, Lianghua Huang, Yu Liu, Yujun Shen, Deli Zhao, and Hengshuang Zhao.
\newblock Anydoor: Zero-shot object-level image customization.
\newblock \emph{arXiv preprint arXiv:2307.09481}, 2023{\natexlab{d}}.

\bibitem[Cho et~al.(2021)Cho, Hong, Jeon, Lee, Sohn, and Kim]{cho2021cats}
Seokju Cho, Sunghwan Hong, Sangryul Jeon, Yunsung Lee, Kwanghoon Sohn, and Seungryong Kim.
\newblock Cats: Cost aggregation transformers for visual correspondence.
\newblock \emph{Advances in Neural Information Processing Systems}, 34:\penalty0 9011--9023, 2021.

\bibitem[Cho et~al.(2022)Cho, Hong, and Kim]{cho2022cats++}
Seokju Cho, Sunghwan Hong, and Seungryong Kim.
\newblock Cats++: Boosting cost aggregation with convolutions and transformers.
\newblock \emph{IEEE Transactions on Pattern Analysis and Machine Intelligence}, 2022.

\bibitem[Dong et~al.(2022)Dong, Wei, and Lin]{dong2022dreamartist}
Ziyi Dong, Pengxu Wei, and Liang Lin.
\newblock Dreamartist: Towards controllable one-shot text-to-image generation via contrastive prompt-tuning.
\newblock \emph{arXiv preprint arXiv:2211.11337}, 2022.

\bibitem[Dosovitskiy et~al.(2020)Dosovitskiy, Beyer, Kolesnikov, Weissenborn, Zhai, Unterthiner, Dehghani, Minderer, Heigold, Gelly, et~al.]{dosovitskiy2020image}
Alexey Dosovitskiy, Lucas Beyer, Alexander Kolesnikov, Dirk Weissenborn, Xiaohua Zhai, Thomas Unterthiner, Mostafa Dehghani, Matthias Minderer, Georg Heigold, Sylvain Gelly, et~al.
\newblock An image is worth 16x16 words: Transformers for image recognition at scale.
\newblock \emph{arXiv preprint arXiv:2010.11929}, 2020.

\bibitem[Epstein et~al.(2023)Epstein, Jabri, Poole, Efros, and Holynski]{epstein2023diffusion}
Dave Epstein, Allan Jabri, Ben Poole, Alexei~A Efros, and Aleksander Holynski.
\newblock Diffusion self-guidance for controllable image generation.
\newblock \emph{arXiv preprint arXiv:2306.00986}, 2023.

\bibitem[Gal et~al.(2022)Gal, Alaluf, Atzmon, Patashnik, Bermano, Chechik, and Cohen-Or]{gal2022image}
Rinon Gal, Yuval Alaluf, Yuval Atzmon, Or Patashnik, Amit~H Bermano, Gal Chechik, and Daniel Cohen-Or.
\newblock An image is worth one word: Personalizing text-to-image generation using textual inversion.
\newblock \emph{arXiv preprint arXiv:2208.01618}, 2022.

\bibitem[Gal et~al.(2023)Gal, Arar, Atzmon, Bermano, Chechik, and Cohen-Or]{gal2023designing}
Rinon Gal, Moab Arar, Yuval Atzmon, Amit~H Bermano, Gal Chechik, and Daniel Cohen-Or.
\newblock Designing an encoder for fast personalization of text-to-image models.
\newblock \emph{arXiv preprint arXiv:2302.12228}, 2023.

\bibitem[Ghosh et~al.(2023)Ghosh, Hajishirzi, and Schmidt]{ghosh2023geneval}
Dhruba Ghosh, Hanna Hajishirzi, and Ludwig Schmidt.
\newblock Geneval: An object-focused framework for evaluating text-to-image alignment.
\newblock \emph{arXiv preprint arXiv:2310.11513}, 2023.

\bibitem[Gu et~al.(2023)Gu, Wang, Zhao, Fu, Xiong, Liu, Zhang, Zhang, Zhang, Jung, et~al.]{gu2023photoswap}
Jing Gu, Yilin Wang, Nanxuan Zhao, Tsu-Jui Fu, Wei Xiong, Qing Liu, Zhifei Zhang, He Zhang, Jianming Zhang, HyunJoon Jung, et~al.
\newblock Photoswap: Personalized subject swapping in images.
\newblock \emph{arXiv preprint arXiv:2305.18286}, 2023.

\bibitem[Han et~al.(2023)Han, Li, Zhang, Milanfar, Metaxas, and Yang]{han2023svdiff}
Ligong Han, Yinxiao Li, Han Zhang, Peyman Milanfar, Dimitris Metaxas, and Feng Yang.
\newblock Svdiff: Compact parameter space for diffusion fine-tuning.
\newblock \emph{arXiv preprint arXiv:2303.11305}, 2023.

\bibitem[Hao et~al.(2023)Hao, Han, Zhao, and Wong]{hao2023vico}
Shaozhe Hao, Kai Han, Shihao Zhao, and Kwan-Yee~K Wong.
\newblock Vico: Detail-preserving visual condition for personalized text-to-image generation.
\newblock \emph{arXiv preprint arXiv:2306.00971}, 2023.

\bibitem[He et~al.(2016)He, Zhang, Ren, and Sun]{he2016deep}
Kaiming He, Xiangyu Zhang, Shaoqing Ren, and Jian Sun.
\newblock Deep residual learning for image recognition.
\newblock In \emph{Proceedings of the IEEE conference on computer vision and pattern recognition}, pages 770--778, 2016.

\bibitem[Hertz et~al.(2022)Hertz, Mokady, Tenenbaum, Aberman, Pritch, and Cohen-Or]{hertz2022prompt}
Amir Hertz, Ron Mokady, Jay Tenenbaum, Kfir Aberman, Yael Pritch, and Daniel Cohen-Or.
\newblock Prompt-to-prompt image editing with cross attention control.
\newblock \emph{arXiv preprint arXiv:2208.01626}, 2022.

\bibitem[Ho et~al.(2020)Ho, Jain, and Abbeel]{ho2020denoising}
Jonathan Ho, Ajay Jain, and Pieter Abbeel.
\newblock Denoising diffusion probabilistic models.
\newblock \emph{Advances in neural information processing systems}, 33:\penalty0 6840--6851, 2020.

\bibitem[Hong et~al.(2022)Hong, Nam, Cho, Hong, Jeon, Min, and Kim]{hong2022neural}
Sunghwan Hong, Jisu Nam, Seokju Cho, Susung Hong, Sangryul Jeon, Dongbo Min, and Seungryong Kim.
\newblock Neural matching fields: Implicit representation of matching fields for visual correspondence.
\newblock \emph{Advances in Neural Information Processing Systems}, 35:\penalty0 13512--13526, 2022.

\bibitem[Horn(1990)]{horn1990hadamard}
Roger~A Horn.
\newblock The hadamard product.
\newblock In \emph{Proc. Symp. Appl. Math}, pages 87--169, 1990.

\bibitem[Huang et~al.(2023)Huang, Liu, Qin, and Chen]{huang2023kv}
Jiancheng Huang, Yifan Liu, Jin Qin, and Shifeng Chen.
\newblock Kv inversion: Kv embeddings learning for text-conditioned real image action editing.
\newblock \emph{arXiv preprint arXiv:2309.16608}, 2023.

\bibitem[Jia et~al.(2023)Jia, Zhao, Chan, Li, Zhang, Gong, Hou, Wang, and Su]{jia2023taming}
Xuhui Jia, Yang Zhao, Kelvin~CK Chan, Yandong Li, Han Zhang, Boqing Gong, Tingbo Hou, Huisheng Wang, and Yu-Chuan Su.
\newblock Taming encoder for zero fine-tuning image customization with text-to-image diffusion models.
\newblock \emph{arXiv preprint arXiv:2304.02642}, 2023.

\bibitem[Jiang et~al.(2021)Jiang, Trulls, Hosang, Tagliasacchi, and Yi]{jiang2021cotr}
Wei Jiang, Eduard Trulls, Jan Hosang, Andrea Tagliasacchi, and Kwang~Moo Yi.
\newblock Cotr: Correspondence transformer for matching across images.
\newblock In \emph{Proceedings of the IEEE/CVF International Conference on Computer Vision}, pages 6207--6217, 2021.

\bibitem[Khandelwal(2023)]{khandelwal2023infusion}
Anant Khandelwal.
\newblock Infusion: Inject and attention fusion for multi concept zero-shot text-based video editing.
\newblock In \emph{Proceedings of the IEEE/CVF International Conference on Computer Vision}, pages 3017--3026, 2023.

\bibitem[Kumari et~al.(2023)Kumari, Zhang, Zhang, Shechtman, and Zhu]{kumari2023multi}
Nupur Kumari, Bingliang Zhang, Richard Zhang, Eli Shechtman, and Jun-Yan Zhu.
\newblock Multi-concept customization of text-to-image diffusion.
\newblock In \emph{Proceedings of the IEEE/CVF Conference on Computer Vision and Pattern Recognition}, pages 1931--1941, 2023.

\bibitem[Lee et~al.(2019)Lee, Cho, and Kiela]{lee2019countering}
Jason Lee, Kyunghyun Cho, and Douwe Kiela.
\newblock Countering language drift via visual grounding.
\newblock \emph{arXiv preprint arXiv:1909.04499}, 2019.

\bibitem[Li et~al.(2023)Li, Li, and Hoi]{li2023blip}
Dongxu Li, Junnan Li, and Steven~CH Hoi.
\newblock Blip-diffusion: Pre-trained subject representation for controllable text-to-image generation and editing.
\newblock \emph{arXiv preprint arXiv:2305.14720}, 2023.

\bibitem[Liu et~al.(2023)Liu, Zhang, Shen, Zheng, Zhu, Feng, Liu, Zhao, Zhou, and Cao]{liu2023cones}
Zhiheng Liu, Yifei Zhang, Yujun Shen, Kecheng Zheng, Kai Zhu, Ruili Feng, Yu Liu, Deli Zhao, Jingren Zhou, and Yang Cao.
\newblock Cones 2: Customizable image synthesis with multiple subjects.
\newblock \emph{arXiv preprint arXiv:2305.19327}, 2023.

\bibitem[Lu et~al.(2020)Lu, Singhal, Strub, Courville, and Pietquin]{lu2020countering}
Yuchen Lu, Soumye Singhal, Florian Strub, Aaron Courville, and Olivier Pietquin.
\newblock Countering language drift with seeded iterated learning.
\newblock In \emph{International Conference on Machine Learning}, pages 6437--6447. PMLR, 2020.

\bibitem[Mou et~al.(2023)Mou, Wang, Song, Shan, and Zhang]{mou2023dragondiffusion}
Chong Mou, Xintao Wang, Jiechong Song, Ying Shan, and Jian Zhang.
\newblock Dragondiffusion: Enabling drag-style manipulation on diffusion models.
\newblock \emph{arXiv preprint arXiv:2307.02421}, 2023.

\bibitem[Nam et~al.(2023)Nam, Lee, Kim, Kim, Cho, Kim, and Kim]{nam2023diffmatch}
Jisu Nam, Gyuseong Lee, Sunwoo Kim, Hyeonsu Kim, Hyoungwon Cho, Seyeon Kim, and Seungryong Kim.
\newblock Diffmatch: Diffusion model for dense matching.
\newblock \emph{arXiv preprint arXiv:2305.19094}, 2023.

\bibitem[OpenAI(2023)]{openai2023gpt4}
OpenAI.
\newblock Gpt-4 technical report, 2023.

\bibitem[Oquab et~al.(2023)Oquab, Darcet, Moutakanni, Vo, Szafraniec, Khalidov, Fernandez, Haziza, Massa, El-Nouby, et~al.]{oquab2023dinov2}
Maxime Oquab, Timoth{\'e}e Darcet, Th{\'e}o Moutakanni, Huy Vo, Marc Szafraniec, Vasil Khalidov, Pierre Fernandez, Daniel Haziza, Francisco Massa, Alaaeldin El-Nouby, et~al.
\newblock Dinov2: Learning robust visual features without supervision.
\newblock \emph{arXiv preprint arXiv:2304.07193}, 2023.

\bibitem[Pearson(1901)]{pearson1901liii}
Karl Pearson.
\newblock Liii. on lines and planes of closest fit to systems of points in space.
\newblock \emph{The London, Edinburgh, and Dublin philosophical magazine and journal of science}, 2\penalty0 (11):\penalty0 559--572, 1901.

\bibitem[Radford et~al.(2021)Radford, Kim, Hallacy, Ramesh, Goh, Agarwal, Sastry, Askell, Mishkin, Clark, et~al.]{radford2021learning}
Alec Radford, Jong~Wook Kim, Chris Hallacy, Aditya Ramesh, Gabriel Goh, Sandhini Agarwal, Girish Sastry, Amanda Askell, Pamela Mishkin, Jack Clark, et~al.
\newblock Learning transferable visual models from natural language supervision.
\newblock In \emph{International conference on machine learning}, pages 8748--8763. PMLR, 2021.

\bibitem[Rombach et~al.(2022)Rombach, Blattmann, Lorenz, Esser, and Ommer]{rombach2022high}
Robin Rombach, Andreas Blattmann, Dominik Lorenz, Patrick Esser, and Bj{\"o}rn Ommer.
\newblock High-resolution image synthesis with latent diffusion models.
\newblock In \emph{Proceedings of the IEEE/CVF conference on computer vision and pattern recognition}, pages 10684--10695, 2022.

\bibitem[Ruiz et~al.(2023{\natexlab{a}})Ruiz, Li, Jampani, Pritch, Rubinstein, and Aberman]{ruiz2023dreambooth}
Nataniel Ruiz, Yuanzhen Li, Varun Jampani, Yael Pritch, Michael Rubinstein, and Kfir Aberman.
\newblock Dreambooth: Fine tuning text-to-image diffusion models for subject-driven generation.
\newblock In \emph{Proceedings of the IEEE/CVF Conference on Computer Vision and Pattern Recognition}, pages 22500--22510, 2023{\natexlab{a}}.

\bibitem[Ruiz et~al.(2023{\natexlab{b}})Ruiz, Li, Jampani, Wei, Hou, Pritch, Wadhwa, Rubinstein, and Aberman]{ruiz2023hyperdreambooth}
Nataniel Ruiz, Yuanzhen Li, Varun Jampani, Wei Wei, Tingbo Hou, Yael Pritch, Neal Wadhwa, Michael Rubinstein, and Kfir Aberman.
\newblock Hyperdreambooth: Hypernetworks for fast personalization of text-to-image models.
\newblock \emph{arXiv preprint arXiv:2307.06949}, 2023{\natexlab{b}}.

\bibitem[Schuhmann et~al.(2021)Schuhmann, Vencu, Beaumont, Kaczmarczyk, Mullis, Katta, Coombes, Jitsev, and Komatsuzaki]{schuhmann2021laion}
Christoph Schuhmann, Richard Vencu, Romain Beaumont, Robert Kaczmarczyk, Clayton Mullis, Aarush Katta, Theo Coombes, Jenia Jitsev, and Aran Komatsuzaki.
\newblock Laion-400m: Open dataset of clip-filtered 400 million image-text pairs.
\newblock \emph{arXiv preprint arXiv:2111.02114}, 2021.

\bibitem[Seo et~al.(2023)Seo, Lee, Cho, Lee, and Kim]{seo2023midms}
Junyoung Seo, Gyuseong Lee, Seokju Cho, Jiyoung Lee, and Seungryong Kim.
\newblock Midms: Matching interleaved diffusion models for exemplar-based image translation.
\newblock In \emph{Proceedings of the AAAI Conference on Artificial Intelligence}, pages 2191--2199, 2023.

\bibitem[Shi et~al.(2023)Shi, Xiong, Lin, and Jung]{shi2023instantbooth}
Jing Shi, Wei Xiong, Zhe Lin, and Hyun~Joon Jung.
\newblock Instantbooth: Personalized text-to-image generation without test-time finetuning.
\newblock \emph{arXiv preprint arXiv:2304.03411}, 2023.

\bibitem[Si et~al.(2023)Si, Huang, Jiang, and Liu]{si2023freeu}
Chenyang Si, Ziqi Huang, Yuming Jiang, and Ziwei Liu.
\newblock Freeu: Free lunch in diffusion u-net.
\newblock \emph{arXiv preprint arXiv:2309.11497}, 2023.

\bibitem[Song et~al.(2020{\natexlab{a}})Song, Meng, and Ermon]{song2020denoising}
Jiaming Song, Chenlin Meng, and Stefano Ermon.
\newblock Denoising diffusion implicit models.
\newblock \emph{arXiv preprint arXiv:2010.02502}, 2020{\natexlab{a}}.

\bibitem[Song and Ermon(2019)]{song2019generative}
Yang Song and Stefano Ermon.
\newblock Generative modeling by estimating gradients of the data distribution.
\newblock \emph{Advances in neural information processing systems}, 32, 2019.

\bibitem[Song et~al.(2020{\natexlab{b}})Song, Sohl-Dickstein, Kingma, Kumar, Ermon, and Poole]{song2020score}
Yang Song, Jascha Sohl-Dickstein, Diederik~P Kingma, Abhishek Kumar, Stefano Ermon, and Ben Poole.
\newblock Score-based generative modeling through stochastic differential equations.
\newblock \emph{arXiv preprint arXiv:2011.13456}, 2020{\natexlab{b}}.

\bibitem[Su et~al.(2023)Su, Chan, Li, Zhao, Zhang, Gong, Wang, and Jia]{su2023identity}
Yu-Chuan Su, Kelvin~CK Chan, Yandong Li, Yang Zhao, Han Zhang, Boqing Gong, Huisheng Wang, and Xuhui Jia.
\newblock Identity encoder for personalized diffusion.
\newblock \emph{arXiv preprint arXiv:2304.07429}, 2023.

\bibitem[Tang et~al.(2023)Tang, Jia, Wang, Phoo, and Hariharan]{tang2023emergent}
Luming Tang, Menglin Jia, Qianqian Wang, Cheng~Perng Phoo, and Bharath Hariharan.
\newblock Emergent correspondence from image diffusion.
\newblock \emph{arXiv preprint arXiv:2306.03881}, 2023.

\bibitem[Tewel et~al.(2023)Tewel, Gal, Chechik, and Atzmon]{tewel2023key}
Yoad Tewel, Rinon Gal, Gal Chechik, and Yuval Atzmon.
\newblock Key-locked rank one editing for text-to-image personalization.
\newblock In \emph{ACM SIGGRAPH 2023 Conference Proceedings}, pages 1--11, 2023.

\bibitem[Truong et~al.(2020{\natexlab{a}})Truong, Danelljan, Gool, and Timofte]{truong2020gocor}
Prune Truong, Martin Danelljan, Luc~V Gool, and Radu Timofte.
\newblock Gocor: Bringing globally optimized correspondence volumes into your neural network.
\newblock \emph{Advances in Neural Information Processing Systems}, 33:\penalty0 14278--14290, 2020{\natexlab{a}}.

\bibitem[Truong et~al.(2020{\natexlab{b}})Truong, Danelljan, and Timofte]{truong2020glu}
Prune Truong, Martin Danelljan, and Radu Timofte.
\newblock Glu-net: Global-local universal network for dense flow and correspondences.
\newblock In \emph{Proceedings of the IEEE/CVF conference on computer vision and pattern recognition}, pages 6258--6268, 2020{\natexlab{b}}.

\bibitem[Truong et~al.(2021)Truong, Danelljan, Yu, and Van~Gool]{truong2021warp}
Prune Truong, Martin Danelljan, Fisher Yu, and Luc Van~Gool.
\newblock Warp consistency for unsupervised learning of dense correspondences.
\newblock In \emph{Proceedings of the IEEE/CVF International Conference on Computer Vision}, pages 10346--10356, 2021.

\bibitem[Truong et~al.(2023)Truong, Danelljan, Timofte, and Van~Gool]{truong2023pdc}
Prune Truong, Martin Danelljan, Radu Timofte, and Luc Van~Gool.
\newblock Pdc-net+: Enhanced probabilistic dense correspondence network.
\newblock \emph{IEEE Transactions on Pattern Analysis and Machine Intelligence}, 2023.

\bibitem[Tumanyan et~al.(2023)Tumanyan, Geyer, Bagon, and Dekel]{tumanyan2023plug}
Narek Tumanyan, Michal Geyer, Shai Bagon, and Tali Dekel.
\newblock Plug-and-play diffusion features for text-driven image-to-image translation.
\newblock In \emph{Proceedings of the IEEE/CVF Conference on Computer Vision and Pattern Recognition}, pages 1921--1930, 2023.

\bibitem[Vaswani et~al.(2017)Vaswani, Shazeer, Parmar, Uszkoreit, Jones, Gomez, Kaiser, and Polosukhin]{vaswani2017attention}
Ashish Vaswani, Noam Shazeer, Niki Parmar, Jakob Uszkoreit, Llion Jones, Aidan~N Gomez, {\L}ukasz Kaiser, and Illia Polosukhin.
\newblock Attention is all you need.
\newblock \emph{Advances in neural information processing systems}, 30, 2017.

\bibitem[Voynov et~al.(2023)Voynov, Chu, Cohen-Or, and Aberman]{voynov2023p+}
Andrey Voynov, Qinghao Chu, Daniel Cohen-Or, and Kfir Aberman.
\newblock $ p+ $: Extended textual conditioning in text-to-image generation.
\newblock \emph{arXiv preprint arXiv:2303.09522}, 2023.

\bibitem[Wei et~al.(2023)Wei, Zhang, Ji, Bai, Zhang, and Zuo]{wei2023elite}
Yuxiang Wei, Yabo Zhang, Zhilong Ji, Jinfeng Bai, Lei Zhang, and Wangmeng Zuo.
\newblock Elite: Encoding visual concepts into textual embeddings for customized text-to-image generation.
\newblock \emph{arXiv preprint arXiv:2302.13848}, 2023.

\bibitem[Xiang et~al.(2023)Xiang, Bao, Li, Su, and Zhu]{xiang2023closer}
Chendong Xiang, Fan Bao, Chongxuan Li, Hang Su, and Jun Zhu.
\newblock A closer look at parameter-efficient tuning in diffusion models.
\newblock \emph{arXiv preprint arXiv:2303.18181}, 2023.

\bibitem[Xiao et~al.(2023)Xiao, Yin, Freeman, Durand, and Han]{xiao2023fastcomposer}
Guangxuan Xiao, Tianwei Yin, William~T Freeman, Fr{\'e}do Durand, and Song Han.
\newblock Fastcomposer: Tuning-free multi-subject image generation with localized attention.
\newblock \emph{arXiv preprint arXiv:2305.10431}, 2023.

\bibitem[Yarom et~al.(2023)Yarom, Bitton, Changpinyo, Aharoni, Herzig, Lang, Ofek, and Szpektor]{yarom2023you}
Michal Yarom, Yonatan Bitton, Soravit Changpinyo, Roee Aharoni, Jonathan Herzig, Oran Lang, Eran Ofek, and Idan Szpektor.
\newblock What you see is what you read? improving text-image alignment evaluation.
\newblock \emph{arXiv preprint arXiv:2305.10400}, 2023.

\bibitem[Zhang et~al.(2023)Zhang, Herrmann, Hur, Cabrera, Jampani, Sun, and Yang]{zhang2023tale}
Junyi Zhang, Charles Herrmann, Junhwa Hur, Luisa~Polania Cabrera, Varun Jampani, Deqing Sun, and Ming-Hsuan Yang.
\newblock A tale of two features: Stable diffusion complements dino for zero-shot semantic correspondence.
\newblock \emph{arXiv preprint arXiv:2305.15347}, 2023.

\bibitem[Zhao et~al.(2023)Zhao, Zheng, Wang, Lan, and Yang]{zhao2023magicfusion}
Jing Zhao, Heliang Zheng, Chaoyue Wang, Long Lan, and Wenjing Yang.
\newblock Magicfusion: Boosting text-to-image generation performance by fusing diffusion models.
\newblock \emph{arXiv preprint arXiv:2303.13126}, 2023.

\end{thebibliography}
}

 \appendix
\renewcommand{\thepage}{A.\arabic{page}}
\renewcommand{\thesection}{\Alph{section}}
\renewcommand{\thesubsection}{\Alph{section}.\arabic{subsection}}
\renewcommand{\thefigure}{A.\arabic{figure}}
\renewcommand{\thetable}{A.\arabic{table}}
\setcounter{page}{1}
\setcounter{figure}{0}
\setcounter{table}{0}

\newpage
\section{Implementation Details}
\label{supp:imp_details}
For all experiments, we used an NVIDIA GeForce RTX 3090 GPU and a DDIM sampler~\cite{song2020denoising}, setting the total sampling time step to $T=50$. We empirically set the time steps to $t\in[4, 50)$ for performing both our appearance matching self-attention and semantic matching guidance. We converted all self-attention modules in every decoder layer $l \in [1,4)$ to the proposed appearance matching self-attention. We chose $\lambda_{c}=0.4$ and $\lambda_{g}=75$ for evaluation on the ViCo~\cite{hao2023vico} dataset, and $\lambda_{c} = 0.4$ and $\lambda_{g} = 50$ for evaluation on the proposed challenging prompt list. 

\section{Dataset}
\label{supp:dataset}
Prior works~\cite{gal2022image,ruiz2023dreambooth,kumari2023multi} in Text-to-Image (T2I) personalization have used different datasets for evaluation. To ensure a fair and unbiased evaluation, ViCo~\cite{hao2023vico} collected an image dataset from these works~\cite{gal2022image,ruiz2023dreambooth,kumari2023multi}, comprising 16 unique concepts, which include 6 toys, 5 live animals, 2 types of accessories, 2 types of containers, and 1 building. For the prompts, ViCo gathered 31 prompts for 11 non-live objects and another 31 prompts for 5 live objects. These were modified from the original DreamBooth~\cite{ruiz2023dreambooth} prompts to evaluate the expressiveness of the objects in more complex textual contexts. For a fair comparison, in this paper, we followed the ViCo dataset and its evaluation settings, producing 8 samples for each object and prompt, totaling 3,969 images.

\begin{figure*}[t]
    \begin{center}
    \includegraphics[width=0.95\textwidth]{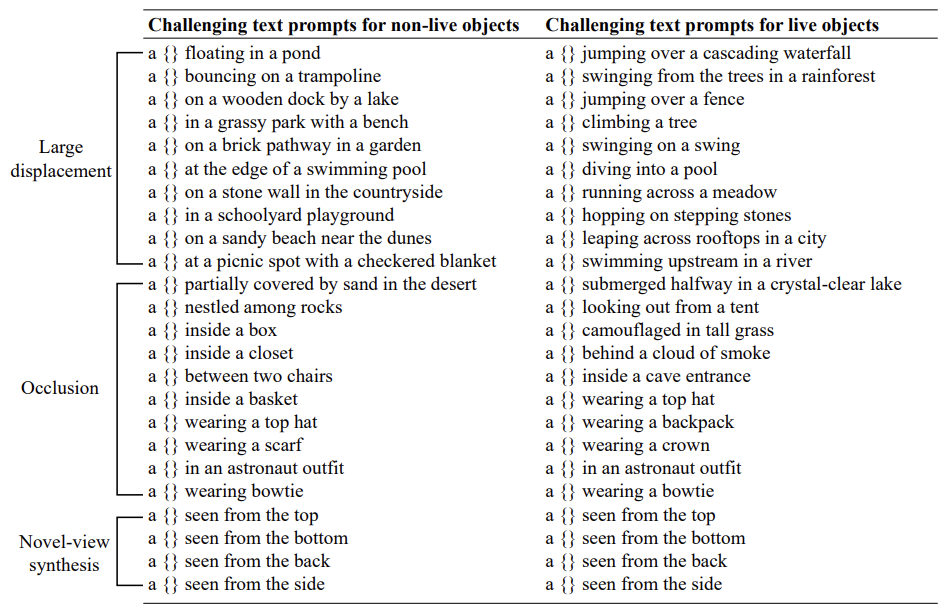} 
    \end{center}
    \vspace{-15pt}
    \caption{\textbf{Challenging text prompt list:} Evaluation prompts in complex, non-rigid scenarios for both non-live and live subjects. `\{\}' represents \(\langle S^* \rangle\) in Textual Inversion~\cite{gal2022image} and `[V] class' in DreamBooth~\cite{ruiz2023dreambooth} and CustomDiffusion~\cite{kumari2023multi}.}

    \vspace{-10pt}
    \label{supp:prompt_list}
\end{figure*}

Our goal is to achieve semantically-consistent T2I personalization in complex non-rigid scenarios. To assess the robustness of our method in intricate settings, we created a prompt dataset using ChatGPT~\cite{openai2023gpt4}, which is categorized into three parts: large displacements, occlusions, and novel-view synthesis. The dataset comprises 10 prompts each for large displacements and occlusions, and 4 for novel-view synthesis, separately for live and non-live objects. Specifically, we define the text-to-image diffusion personalization task, provide an example prompt list from ViCo, and highlight the necessity of a challenging prompt list aligned with the objectives of our research. We then asked ChatGPT to create distinct prompt lists for each category. The resulting prompt list, tailored for complex non-rigid personalization scenarios, is detailed in Figure~\ref{supp:prompt_list}.

\section{Baseline and Comparison}     

\label{supp:baselines}
\subsection{Baseline}
DreamMatcher is designed to be compatible with any T2I personalized model. We implemented our method using three baselines: Textual Inversion~\cite{gal2022image}, DreamBooth~\cite{ruiz2023dreambooth}, and CustomDiffusion~\cite{kumari2023multi}. 

Textual Inversion~\cite{gal2022image} encapsulates a given subject into 768-dimensional textual embeddings derived from the special token $\langle S^* \rangle$. Using a few reference images, this is achieved by training the textual embeddings while keeping the T2I diffusion model frozen. During inference, the model can generate novel renditions of the subject by manipulating the target prompt with $\langle S^* \rangle$. DreamBooth~\cite{ruiz2023dreambooth} extends this approach by further fine-tuning a T2I diffusion model with a unique identifier and the class name of the subject (e.g., \textit{A} [V] \textit{cat}). However, fine-tuning all parameters can lead to a language shift problem~\cite{lu2020countering, lee2019countering}. To address this, DreamBooth proposes a class-specific prior preservation loss, which trains the model with diverse samples generated by pre-trained T2I models using the category name as a prompt (e.g., \textit{A cat}). Lastly, CustomDiffusion~\cite{kumari2023multi} demonstrates that fine-tuning only a subset of parameters, specifically the cross-attention projection layers, is efficient for learning new concepts. Similar to DreamBooth, this is implemented by using a text prompt that combines a unique instance with a general category, and it also includes a regularization dataset from the large-scale open image-text dataset~\cite{schuhmann2021laion}. Despite promising results, the aforementioned approaches frequently struggle to accurately mimic the appearance of the subject, including colors, textures, and shapes. To address this, we propose a tuning-free plug-in method that significantly enhances the reference appearance while preserving the diverse structure from target prompts. 

\subsection{Comparision}
We benchmarked DreamMatcher against previous tuning-free plug-in models, FreeU~\cite{si2023freeu} and MagicFusion~\cite{zhao2023magicfusion}, and also against the optimization-based model, ViCo~\cite{hao2023vico}. 

The key insight of FreeU~\cite{si2023freeu} is that the main backbone of the denoising U-Net contributes to low-frequency semantics, while its skip connections focus on high-frequency details. Leveraging this observation, FreeU proposes a frequency-aware reweighting technique for these two distinct features, and demonstrates improved generation quality when integrated into DreamBooth~\cite{ruiz2023dreambooth}. MagicFusion~\cite{zhao2023magicfusion} introduces a saliency-aware noise blending method, which involves combining the predicted noises from two distinct pre-trained diffusion models. MagicFusion demonstrates its effectiveness in T2I personalization when integrating a personalized model, DreamBooth~\cite{ruiz2023dreambooth}, with a general T2I diffusion model. ViCo~\cite{hao2023vico} optimizes an additional image adapter designed with the concept of key-value replacement.

 \begin{figure}[t]
    \centering
    \includegraphics[width=0.48\textwidth]{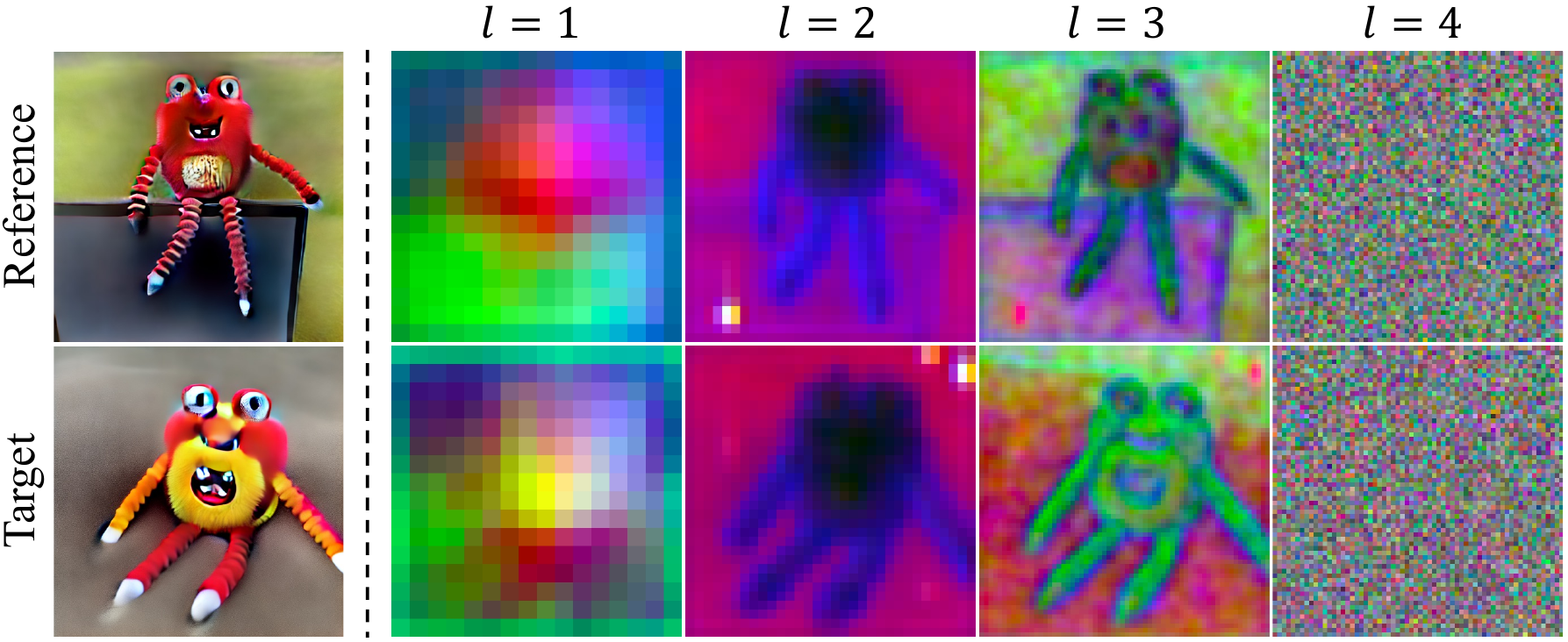}
    \vspace{-15pt}    
    \caption{\textbf{Diffusion feature visualization at different decoder layers:} The left side displays intermediate estimated reference and target images at 50\% of the reverse diffusion process. The target is generated by DreamBooth~\cite{ruiz2023dreambooth} using the prompt \textit{A $\langle S^* \rangle$ on the beach}. The right side visualizes the top three principal components of diffusion feature descriptors from different decoder layers $l$. Semantically similar regions share similar colors.}
    \vspace{-5pt}
    \label{supp:feat_vis}
\end{figure}

\section{Evaluation}
\label{supp:evaluation}
\subsection{Evaluation Metrics}
\label{supp:metrics}
For evaluation, we focused on two primary aspects: subject fidelity and prompt fidelity. For subject fidelity, following prior studies~\cite{hao2023vico, gal2022image, ruiz2023dreambooth, kumari2023multi}, we adopted the CLIP~\cite{radford2021learning} and DINO~\cite{caron2021emerging} image similarity metrics, denoted as $I_\mathrm{CLIP}$ and $I_\mathrm{DINO}$, respectively. Note that $I_\mathrm{DINO}$ is our preferred metric for evaluating subject expressivity. As mentioned in~\cite{ruiz2023dreambooth, hao2023vico}, DINO is trained in a self-supervised manner to distinguish objects within the same category, so that it is more suitable for evaluating different methods that aim to mimic the visual attributes of the same subject.
For prompt fidelity, following~\cite{hao2023vico, gal2022image, ruiz2023dreambooth, kumari2023multi}, we adopted the image-text similarity metric $T_\text{CLIP}$, comparing CLIP visual features of the generated images to CLIP textual features of the corresponding text prompts, excluding placeholders. Following previous works~\cite{hao2023vico, gal2022image, ruiz2023dreambooth, kumari2023multi}, we used ViT-B/32~\cite{dosovitskiy2020image} and ViT-S/16~\cite{dosovitskiy2020image} for CLIP and DINO, respectively.

\subsection{User study}
\label{supp:user_study}
\begin{figure*}[t]
    \begin{center}
    \includegraphics[width=0.60\textwidth]{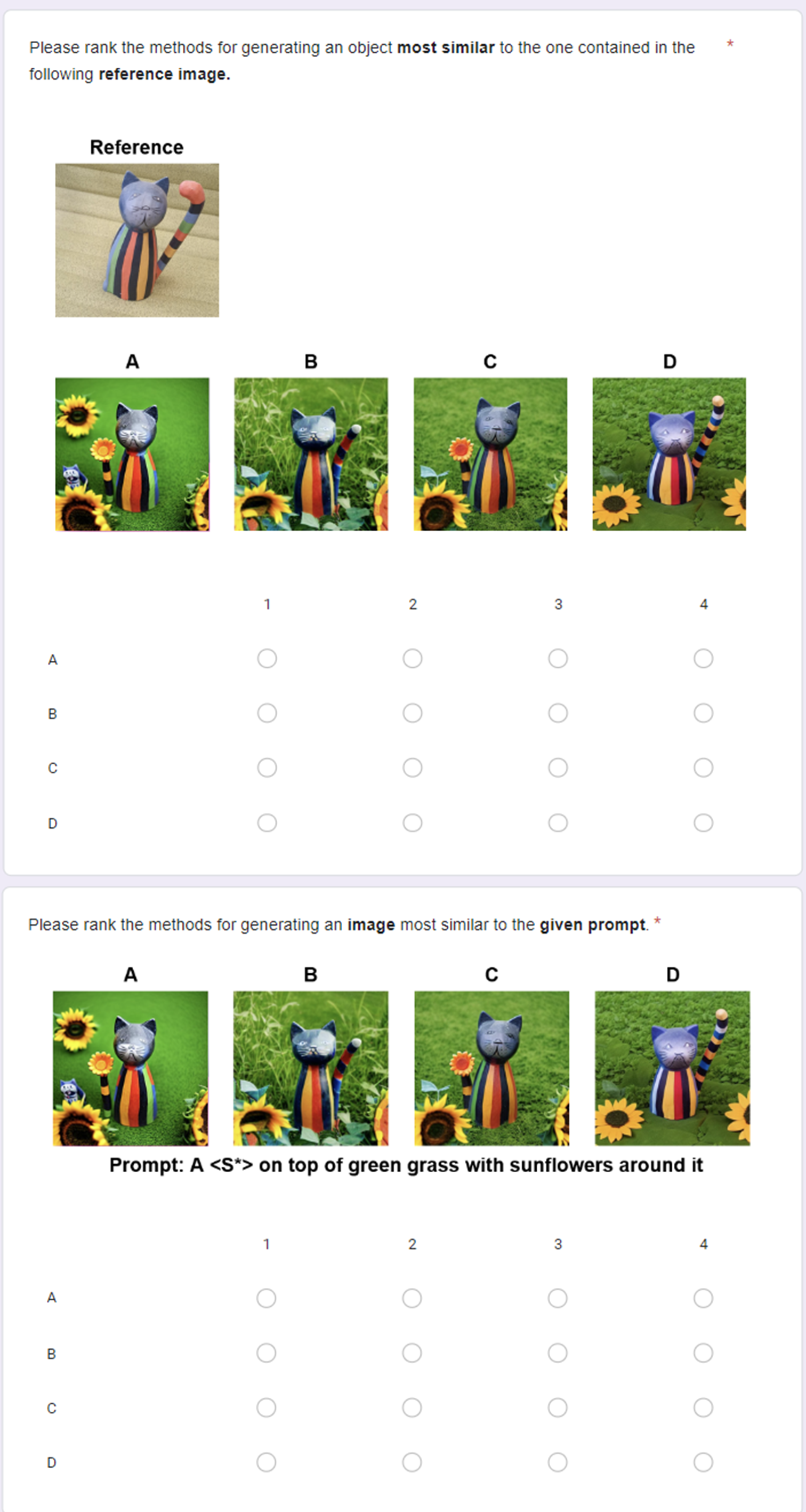} 
    \end{center}
    \caption{\textbf{An example of a user study comparing DreamMatcher with previous methods:} For subject fidelity, we provide the reference image and generated images from different methods, ViCo~\cite{hao2023vico}, FreeU~\cite{si2023freeu}, MagicFusion~\cite{zhao2023magicfusion} and DreamMatcher. For prompt fidelity, we provide the target prompt and the generated images from those methods. For a fair comparison, we randomly choose the image samples from a large, unbiased pool.}

    \vspace{-10pt}
    \label{supp:user_study_format}
\end{figure*}

An example question of the user study is provided in Figure~\ref{supp:user_study_format}. We conducted a paired human preference study about subject and prompt fidelity, comparing DreamMatcher to previous works~\cite{hao2023vico, si2023freeu, zhao2023magicfusion}. The results are summarized in Figure~\ref{qual:user_study} in the main paper. For subject fidelity, participants were presented with a reference image and generated images from different methods, and were asked which better represents the subject in the reference. For prompt fidelity, they were shown the generated images from different works alongside the corresponding text prompt, and were asked which aligns more with the given prompt. 45 users responded to 32 comparative questions, resulting in a total of 1440 responses. We distributed two different questionnaires, with 23 users responding to one and 22 users to the other. Note that samples were chosen randomly from a large, unbiased pool.

\section{Analysis}
\label{supp:analysis}

 \begin{figure}[t]
    \centering
    \includegraphics[width=0.48\textwidth]{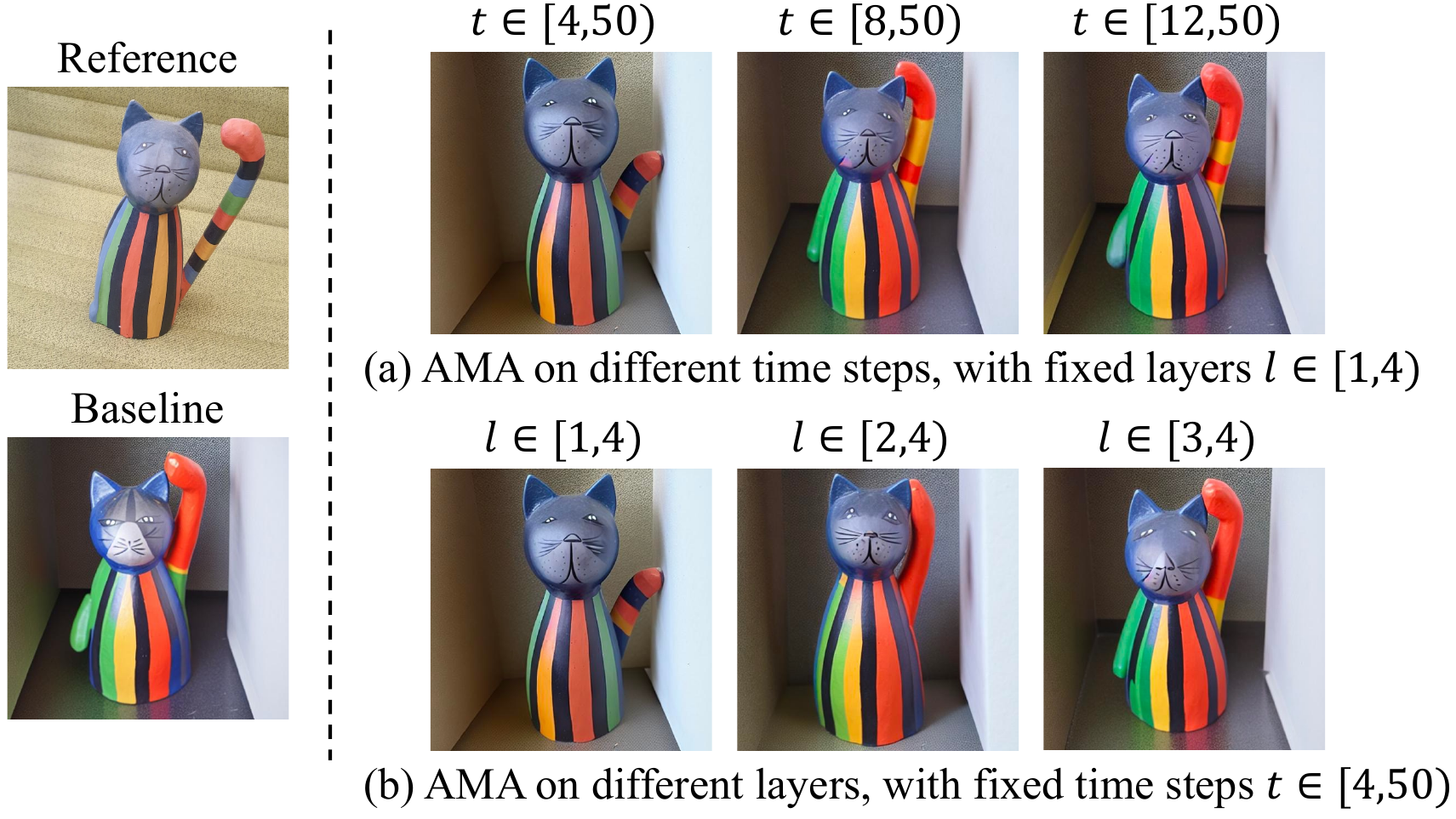}
    \vspace{-20pt}    
    \caption{\textbf{Ablating AMA on different time steps and layers: }The left section shows a reference image and a target image generated by the baseline~\cite{ruiz2023dreambooth}. The right section displays the improved target image generated by appearance matching self-attention on (a) different time steps and (b) different decoder layers. For this ablation study, we do not use semantic matching guidance.}
    \vspace{-10pt}
    \label{supp:ablation_ama}
\end{figure}

\subsection{Appearance Matching Self-Attention}
\label{supp:ama}
\paragrapht{Feature extraction.} Figure~\ref{supp:feat_vis} visualizes PCA~\cite{pearson1901liii} results on feature descriptors extracted from different decoder layers. Note that, for this analysis, we do not apply any of our proposed techniques. PCA is applied to the intermediate feature descriptors of the estimated reference image and target image from DreamBooth~\cite{ruiz2023dreambooth}, at 50\% of the reverse diffusion process. Our primary insight is that earlier layers capture high-level semantics, while later layers focus on finer details of the generated images. Specifically, $l=1$ captures overly high-level and low-resolution semantics, failing to provide sufficient semantics for finding correspondence. Conversely, $l=4$ focuses on too fine-grained details, making it difficult to find semantically-consistent regions between features. In contrast, $l=2$ and $l=3$ strike a balance, focusing on sufficient semantical and structural information to facilitate semantic matching. Based on this analysis, we use concatenated feature descriptors from decoder layers $l \in [2,3]$, resulting in $\psi_{t} \in \mathbb{R}^{H \times W\times 1920}$. We then apply PCA to these feature descriptors, which results in $\psi_{t} \in \mathbb{R}^{H \times W\times 256}$ to enhance matching accuracy and reduce memory consumption. The diffusion feature visualization across different time steps is presented in Figure~\ref{qual:feat_vis} in our main paper.

Note that our approach differs from prior works~\cite{tang2023emergent,zhang2023tale,mou2023dragondiffusion}, which select a specific time step and inject the corresponding noise into clean RGB images before passing them through the pre-trained diffusion model. In contrast, we utilize diffusion features from each time step of the reverse diffusion process to find semantic matching during each step of the personalization procedure.

\vspace{3pt}
\paragrapht{AMA on different time steps and layers.} We ablate starting time steps and decoder layers in relation to the proposed appearance matching self-attention (AMA) module. Figure~\ref{supp:ablation_ama} summarizes the results. Interestingly, we observe that applying AMA at earlier time steps and decoder layers effectively corrects the overall appearance of the subject, including shapes, textures, and colors. In contrast, AMA applied at later time steps and layers tends to more closely preserve the appearance of the subject as in the baseline. Note that injecting AMA at every time step yields sub-optimal results, as the baselines prior to time step 4 have not yet constructed the target image layout. Based on this analysis, we converted the self-attention module in the pre-trained U-Net into the appearance matching self-attention for $t \in [4,50)$ and $l \in [1,4)$ in all our evaluations.

\begin{figure}[t]
    \centering
    \includegraphics[width=0.48\textwidth]{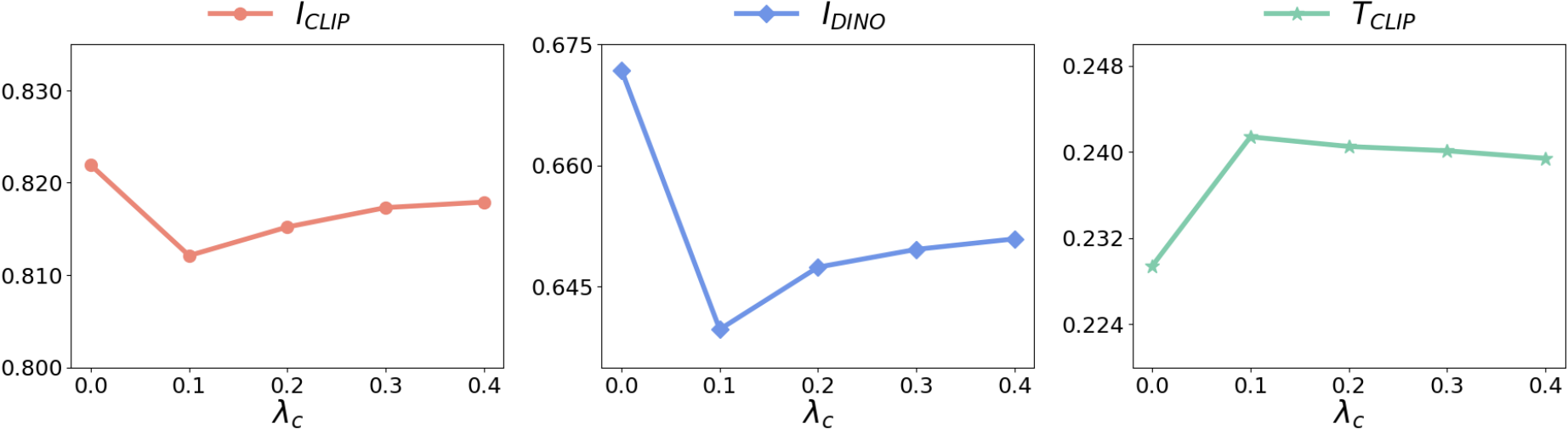}
    \vspace{-20pt}
    \caption{\textbf{Relation between $\lambda_{c}$ and personalization fidelity.} In this ablation study, we evaluate our method using the proposed challenging prompt list.}
    \vspace{-5pt}
    \label{supp:cons}
\end{figure}

\begin{figure}[t]
    \centering
    \includegraphics[width=0.48\textwidth]{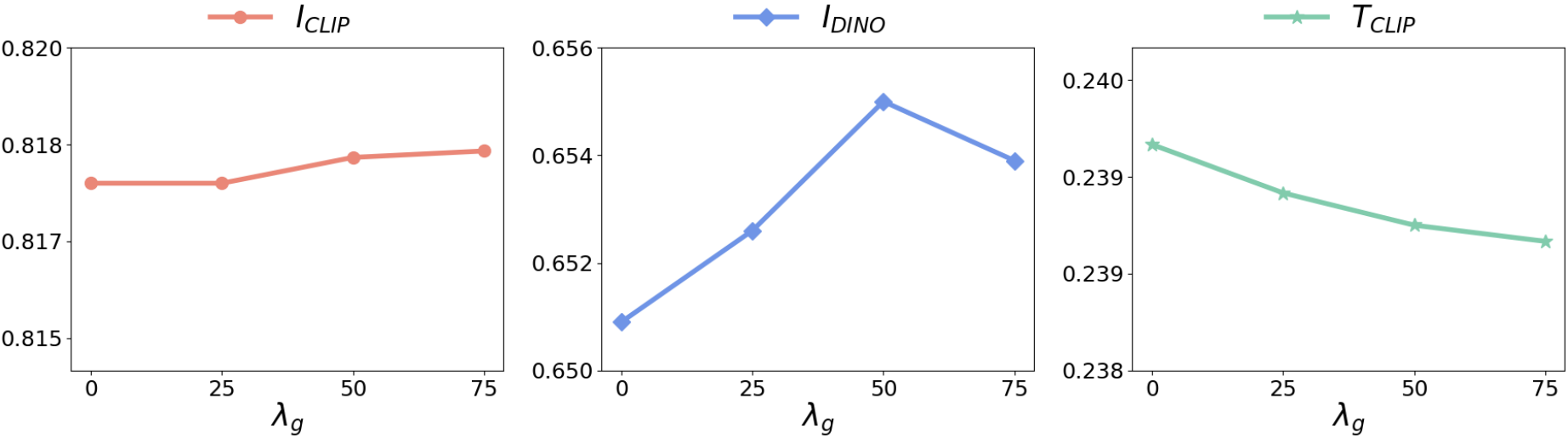}
    \vspace{-20pt}
    \caption{\textbf{Relation between $\lambda_{g}$ and personalization fidelity.} In this ablation study, we evaluate our method using the proposed challenging prompt list.}
    \vspace{-10pt}
    \label{supp:guidance}
\end{figure}

\subsection{Consistency Modeling}
\label{supp:cf}
In Figure~\ref{supp:cons}, we show the quantitative relationship between the cycle-consistency hyperparameter $\lambda_{c}$ and personalization fidelity. As we first introduce $\lambda_{c}$, prompt fidelity $T_\mathrm{CLIP}$ drastically improves, demonstrating that the confidence mask effectively filters out erroneous matches, allowing the model to preserve the detailed target structure. Subsequently, higher $\lambda_{c}$ values inject more reference appearance into the target structure, increasing $I_\mathrm{DINO}$ and $I_\mathrm{CLIP}$, but slightly sacrificing prompt fidelity $T_\mathrm{CLIP}$. This indicates that users can control the extent of reference appearance and target structure preservation by adjusting $\lambda_{c}$. The pseudo code for overall AMA is available in Algorithm~\ref{supp:pseudo1}.

\subsection{Semantic Matching Guidance}
\label{supp:smg}
In Figure~\ref{supp:guidance}, we display the quantitative relationship between semantic matching guidance $\lambda_{g}$ and personalization fidelity. Increasing $\lambda_{g}$ enhances subject fidelity $I_\mathrm{DINO}$ and $I_\mathrm{CLIP}$, by directing the generated target $\hat{z}^\mathrm{Y}_{0,t}$ closer to the clean reference latent $z^\mathrm{X}_0$. However, excessively high $\lambda_{g}$ can reduce subject fidelity due to discrepancies between the reference and target latents in early time steps. We carefully ablated the parameter $\lambda_{g}$ and chose $\lambda_{g} = 75$ for the ViCo dataset and $\lambda_{g} = 50$ for the proposed challenging dataset. The pseudo code for overall semantic matching guidance is available in Algorithm~\ref{supp:pseudo2}.

 \begin{figure}[t]
    \centering
    \includegraphics[width=0.48\textwidth]{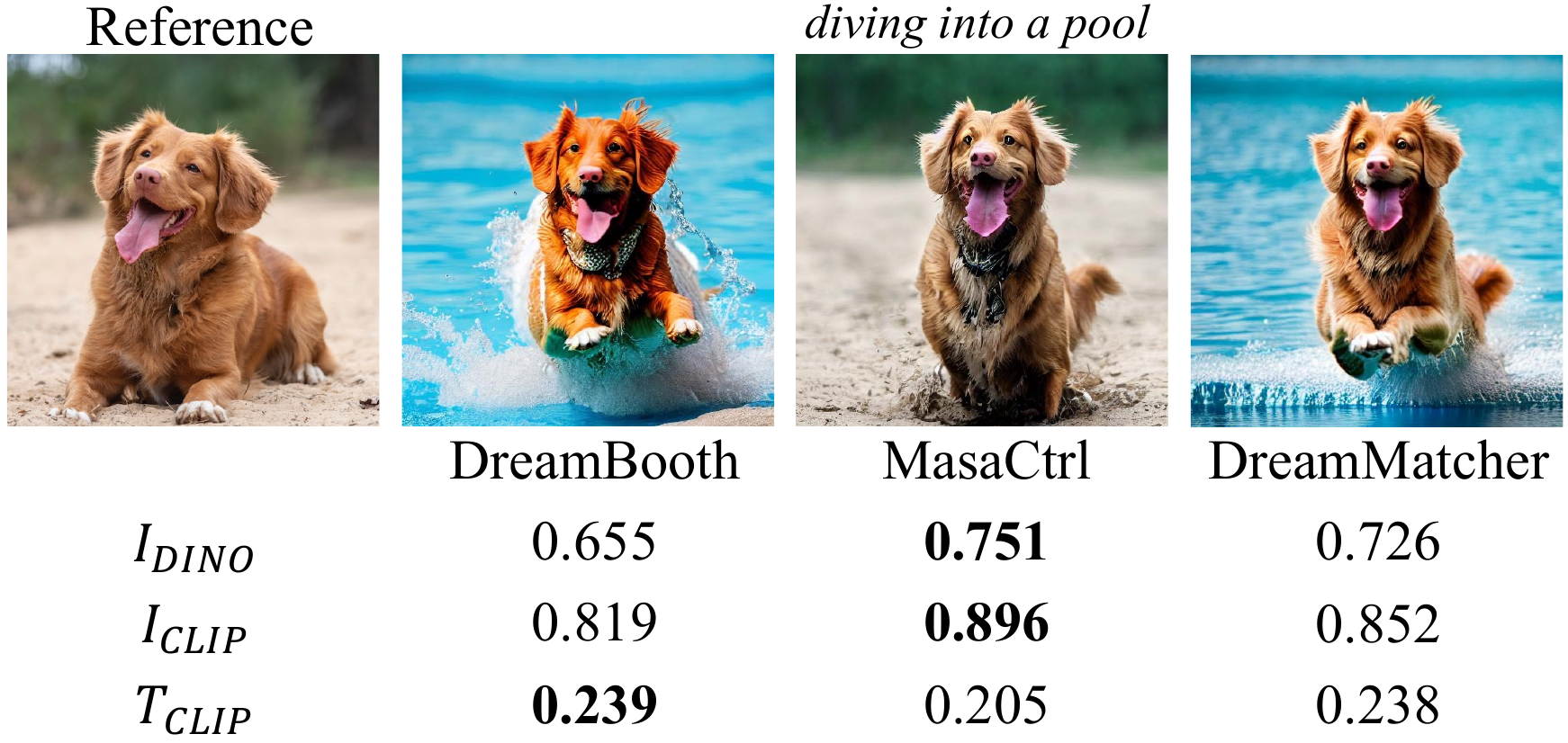}
    \vspace{-20pt}    
    \caption{\textbf{DreamBooth vs. MasaCtrl vs. DreamMatcher. }}
    \vspace{-5pt}
    \label{supp:qual_masactrl}
\end{figure}

\begin{figure}[t]
    \centering
    \includegraphics[width=0.40\textwidth]{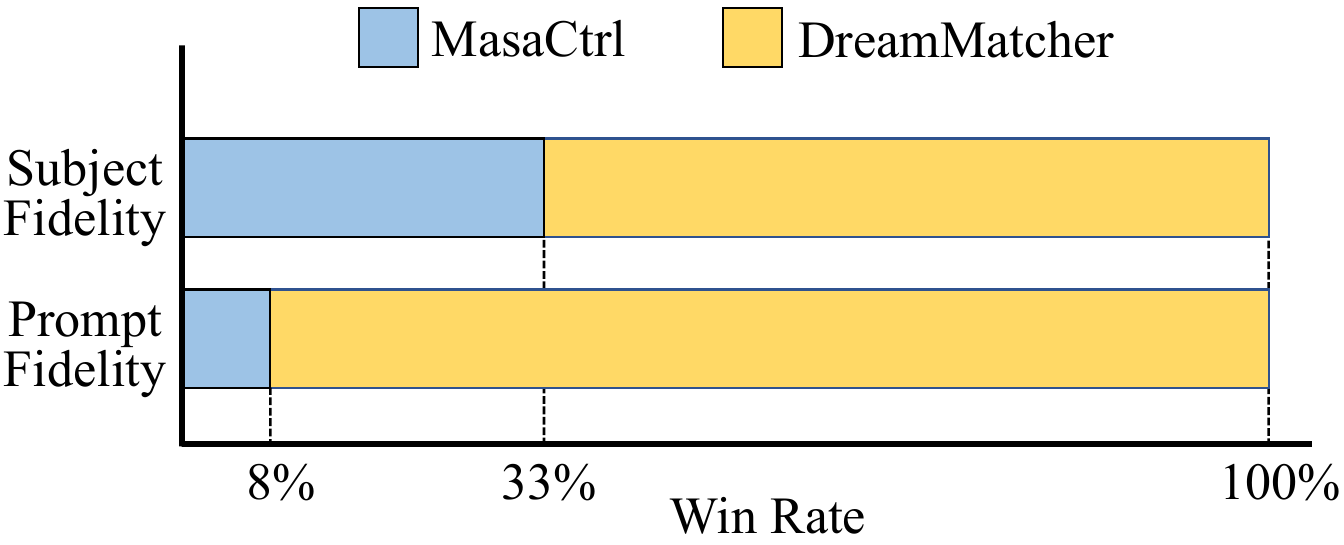}
    \vspace{-10pt}
    \caption{\textbf{User study:} In this study, DreamBooth~\cite{ruiz2023dreambooth} is used as the baseline for both MasaCtrl and DreamMatcher.}
    \vspace{-10pt}
    \label{supp:masa_user}
\end{figure}

\subsection{Key-Value Replacement vs. DreamMatcher} 
MasaCtrl~\cite{cao2023masactrl} introduced a key-value replacement technique for local editing tasks. Several subsequent works~\cite{cao2023masactrl, mou2023dragondiffusion, chen2023anydoor, chen2023fec, huang2023kv, khandelwal2023infusion} have adopted and further developed this framework. As shown in Figure~\ref{supp:qual_masactrl}, which provides a qualitative comparison of DreamMatcher with MasaCtrl, the key-value replacement is prone to producing subject-centric images, often having poses similar to those of the subject in the reference image. This tendency arises because key-value replacement disrupts the target structure from the pre-trained self-attention module and relies on sub-optimal matching between target keys and reference queries. Furthermore, this technique does not consider the uncertainty of predicted matches, which leads to the injection of irrelevant elements from the reference image into the changed background or into newly emergent objects that are produced by the target prompts.

In contrast, DreamMatcher preserves the fixed target structure and accurately aligns the reference appearance by explicitly leveraging semantic matching. Our method also takes into account the uncertainty of predicted matches, thereby filtering out erroneous matches and maintaining newly introduced image elements by the target prompts. Note that the image similarity metrics $I_\mathrm{DINO}$ and $I_\mathrm{CLIP}$ do not simultaneously consider both the preservation of the target structure and the reflection of the reference appearance. They only calculate the similarities between the overall pixels of the reference and generated images. As a result, the key-value replacement, which generates subject-centric images and injects irrelevant elements from reference images into the target context, achieves better image similarities than DreamMatcher, as seen in Table~\ref{quan:comp_analysis} in the main paper. However, as shown in Figure~\ref{supp:qual_masactrl}, DreamMatcher more accurately aligns the reference appearance into the target context, even with large structural displacements. More qualitative comparisons are provided in Figures~\ref{supp:qual_with_sota} and \ref{supp:qual_with_sota2}.

\begin{figure*}[t]
    \begin{center}
    \includegraphics[width=0.60\textwidth]{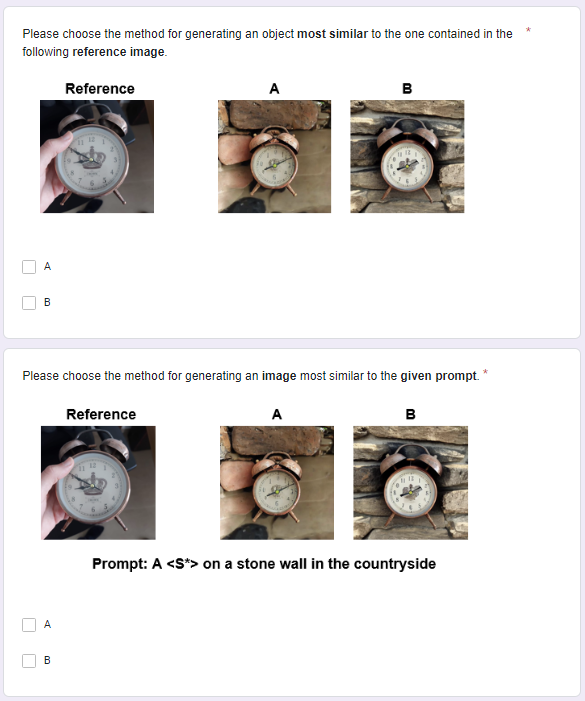} 
    \end{center}
    \caption{\textbf{An example of a user study comparing DreamMatcher with MasaCtrl~\cite{cao2023masactrl}:} For subject fidelity, we provide the reference image and images generated from two different methods, MasaCtrl and DreamMatcher. For prompt fidelity, the target prompt and generated images from these two methods are provided. For a fair comparison, image samples are randomly chosen from a large, unbiased pool.}

    \vspace{-10pt}
    \label{supp:user_study_format_masactrl}
\end{figure*}

This is further demonstrated in a user study comparing MasaCtrl~\cite{cao2023masactrl} and DreamMatcher, summarized in Figure~\ref{supp:masa_user}. A total of 39 users responded to 32 comparative questions, resulting in 1248 responses. These responses were divided between two different questionnaires, with 20 users responding to one and 19 to the other. Samples were chosen randomly from a large, unbiased pool. An example of this user study is shown in Figure~\ref{supp:user_study_format_masactrl}. DreamMatcher significantly surpasses MasaCtrl for both fidelity by a large margin, demonstrating the effectiveness of our proposed matching-aware value injection method.

\begin{table}[t]
\centering
\resizebox{1\columnwidth}{!}{%
\begin{tabular}{l|ccc}
    \toprule
     Method & $I_\mathrm{DINO}$ $\uparrow$ & $I_\mathrm{CLIP}$ $\uparrow$  & $T_\mathrm{CLIP}$ $\uparrow$ \\
    \midrule
    Warping only reference values (DreamMatcher) &\textbf{0.680} &\textbf{0.821} & 0.231\\
    Warping both reference keys and values & 0.654  & 0.809 & \textbf{0.235} \\
    \bottomrule
\end{tabular}%
}
\vspace{-7pt}
\caption{\textbf{Ablation study on key retention.}}
\vspace{-5pt}
\label{quan:key_retention}
\end{table}

\subsection{Justification of Key Retention}
\label{supp:key_retention}
DreamMatcher brings warped reference values to the target structure through semantic matching. This design choice is rational because we leverage the pre-trained U-Net, which has been trained with pairs of queries and keys sharing the same structure. This allows us to preserve the pre-trained target structure path by keeping target keys and queries unchanged. To validate this, Table~\ref{quan:key_retention} shows a quantitative comparison between warping only reference values and warping both reference keys and values, indicating that anchoring the target structure path while only warping the reference appearance is crucial for overall performance. Concerns may arise regarding the misalignment between target keys and warped reference values. However, we emphasize that our appearance matching self-attention accurately aligns reference values with the target structure, ensuring that target keys and warped reference values are geometrically aligned as they were pre-trained.

\begin{figure}[t]
    \centering
    \includegraphics[width=0.48\textwidth]{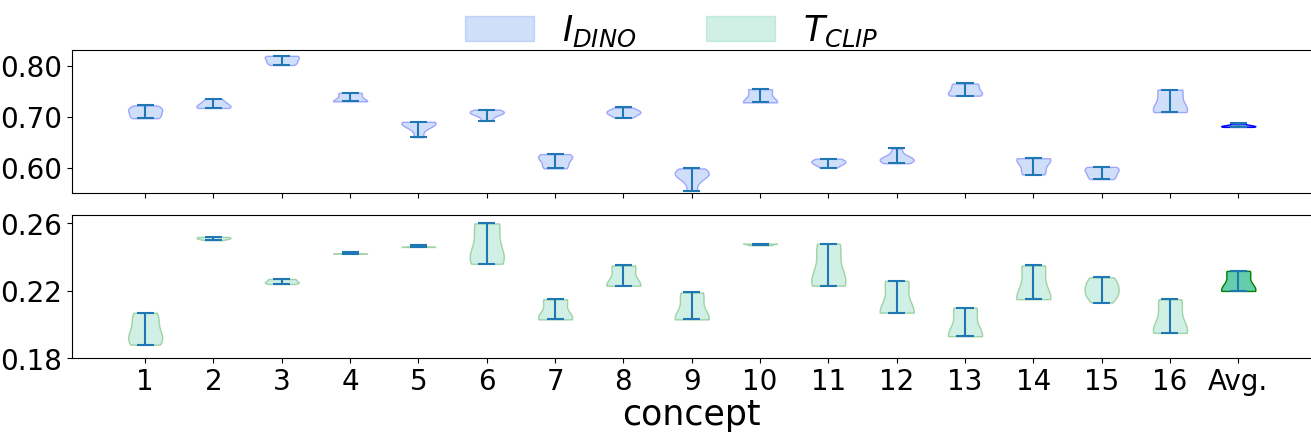}
    \vspace{-20pt}
    \caption{\textbf{Statistical results from 5 sets of randomly selected reference images.}}
    \label{supp:reference}
    \vspace{-10pt}
\end{figure}

\subsection{Reference Selection}
\label{supp:reference_select}
We evaluate the stability of DreamMatcher against variations in reference image selection by measuring the variance of all metrics across five sets of randomly selected reference images. Figure~\ref{supp:reference} indicates that all metrics are closely distributed around the average. Specifically, the average $I_\mathrm{DINO}$ is 0.683 with a variance of $6e{-6}$, and the average $T_\mathrm{CLIP}$ is 0.225 with a variance of $3e{-5}$. This highlights that our method is robust to reference selection and consistently generates reliable results. We further discuss the qualitative comparisions- with different reference images in Section~\ref{sec:limitation}.

\begin{table}[t]
\centering
\resizebox{1\columnwidth}{!}{%
\begin{tabular}{l|ccc}
    \toprule
     Method & $I_\mathrm{DINO}$ $\uparrow$ & $I_\mathrm{CLIP}$ $\uparrow$  & $T_\mathrm{CLIP}$ $\uparrow$ \\
    \midrule
    \midrule
    Textual Inversion~\cite{gal2022image}  &0.529 &0.762 & \textbf{0.220}\\
    \midrule
    Stable Diffusion (SD) & 0.516 &{0.770} & {0.215} \\
    DreamMatcher   &\textbf{0.571} \textcolor{blue}{(+10.74\%)} &\textbf{0.785} \textcolor{blue}{(+1.95\%)}  & 0.214 \textcolor{red}{(-0.50\%)} \\
    \bottomrule
\end{tabular}%
}
\vspace{-7pt}
\caption{\textbf{Quantitative results of DreamMatcher on Stable Diffusion.}}
\vspace{-5pt}
\label{quan:db_on_sd}
\end{table}

\begin{figure}[t]
    \centering
    \includegraphics[width=0.48\textwidth]{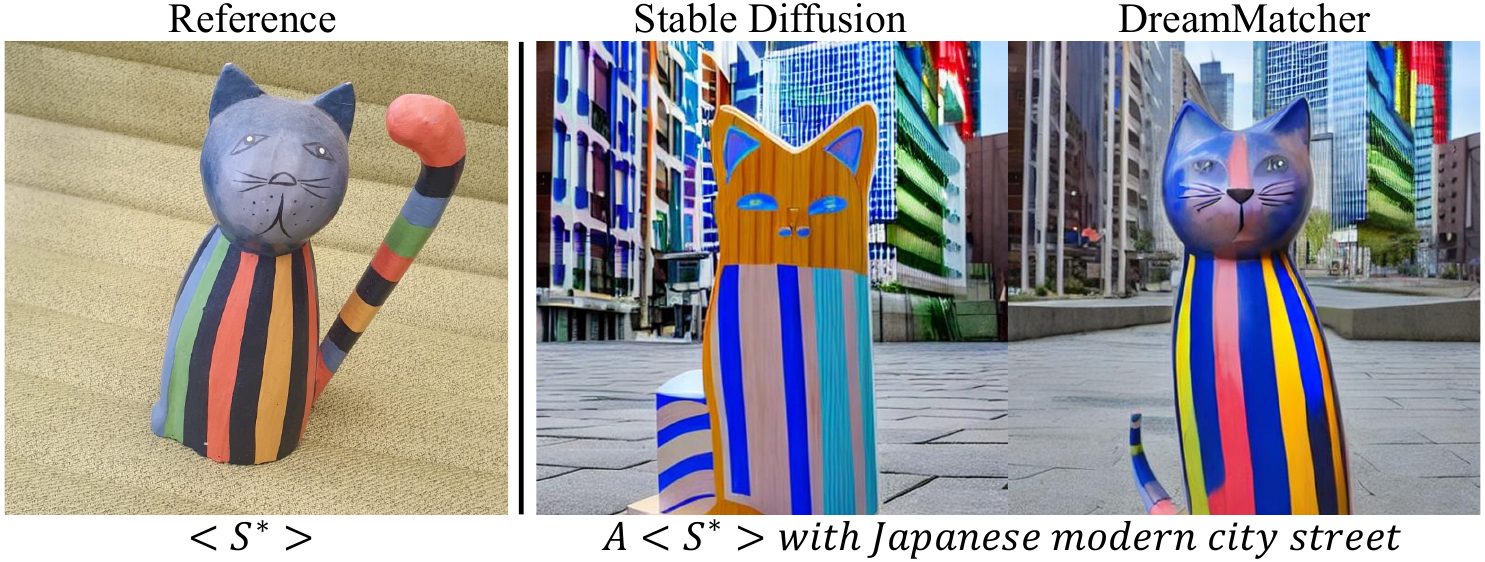}
    \vspace{-20pt}
    \caption{\textbf{Qualitative results of DreamMatcher on Stable Diffusion.}}
    \label{supp:qual_db_on_sd}
    \vspace{-10pt}
\end{figure}

\subsection{DreamMatcher on Stable Diffusion}
\label{supp:base_sd}
DreamMatcher is a plug-in method dependent on the baseline, so we evaluated DreamMatcher on pre-trained personalized models~\citep{gal2022image, ruiz2023dreambooth, kumari2023multi} in the main paper. In this section, we also evaluated DreamMatcher using Stable Diffusion as a baseline. Table~\ref{quan:db_on_sd} and Figure~\ref{supp:qual_db_on_sd} show that DreamMatcher enhances subject fidelity without any off-the-shelf pre-trained models, even surpassing $I_\mathrm{DINO}$ and $I_\mathrm{CLIP}$ of Textual Inversion which optimizes the 769-dimensional text embeddings.

\subsection{Multiple Subjects Personalization}
\label{supp:multiple_ref}
\begin{figure}[t]
    \centering
    \includegraphics[width=0.48\textwidth]{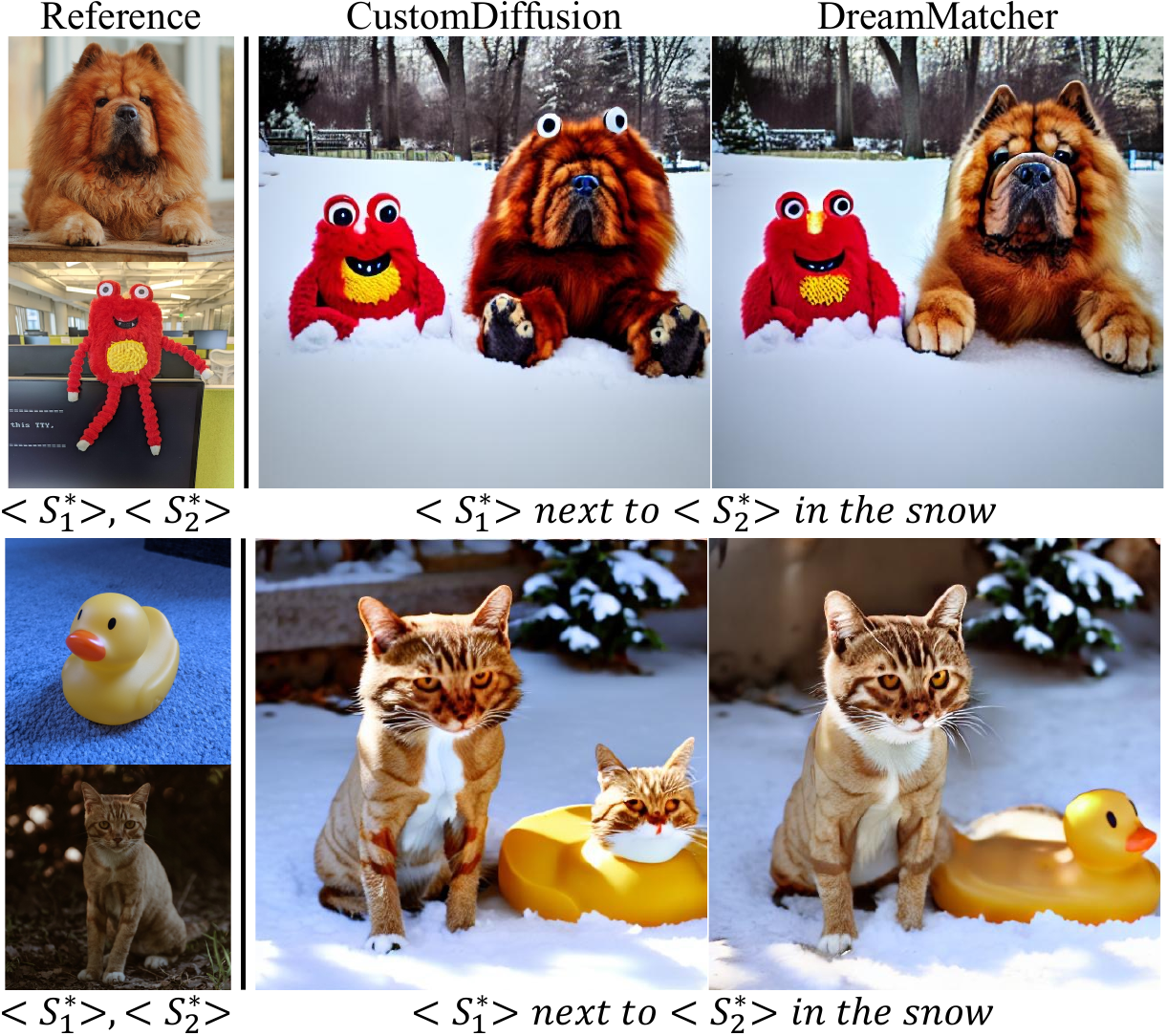}
    \vspace{-20pt}
    \caption{\textbf{Qualitative results of DreamMatcher for multiple subject personalization.}}
    \label{supp:multiple_subj}
    \vspace{-5pt}
\end{figure}

As shown in Figure~\ref{supp:multiple_subj}, we extend DreamMatcher for multiple subjects. For this experiments, we used CustomDiffusion~\citep{kumari2023multi} as a baseline. Note that a simple modification, which involves batching two different subjects as input, enables this functionality.

\subsection{Computational Complexity}
\label{supp:computation} We investigate time and memory consumption on different configurations of our framework, as summarized in Table~\ref{supp:quan_complexity}. As seen, DreamMatcher significantly improves subject appearance with a reasonable increase in time and memory, compared to the baseline DreamBooth~\cite{ruiz2023dreambooth}. Additionally, we observe that reducing the PCA~\cite{pearson1901liii} dimension of feature descriptors before building the cost volume does not affect the overall performance, while dramatically reducing time consumption. Note that our method, unlike previous training-based~\cite{jia2023taming, shi2023instantbooth, chen2023subject, xiao2023fastcomposer, gal2023designing, wei2023elite, chen2023photoverse, su2023identity, li2023blip} or optimization-based approaches~\cite{hao2023vico}, does not require any training or fine-tuning.

    \begin{table}[t]
    \centering
    \resizebox{1
    \columnwidth}{!}{
    \begin{tabular}{c|l|ccccc}
        \toprule
        & Component & $I_\mathrm{DINO}$& $I_\mathrm{CLIP}$ &$T_\mathrm{CLIP}$& Time [s] & Mem. [GB] \\
        \midrule
        (I) & Baseline (DreamBooth~\cite{ruiz2023dreambooth}) &0.638&0.808&0.237&0.27&4.41\\
        \midrule
        (II) & (I) + Appearance Matching Self-Att. (PCA 64) &0.675&0.818&0.233&0.38&4.49\\
        (III) & (I) + Appearance Matching Self-Att. (PCA 256) &0.676&0.818&0.232&0.46&4.56\\
        (IV) & (II) + Semantic Matching Guid.  &0.680&0.820&0.231&0.57&4.85 \\
        \bottomrule
    \end{tabular}}
    \vspace{-7pt}
    \caption{\textbf{Computational complexity:} We used DreamBooth~\cite{ruiz2023dreambooth} as the baseline. For this analysis, we examine the time consumption for a single sampling time step.}
    \vspace{-5pt} 
    \label{supp:quan_complexity}
    \end{table}


\section{More Results}
\label{supp:more_results}
\subsection{Comparison with Baselines}
\label{supp:comp_base}
\begin{figure*}[t]
    \begin{center}
    \includegraphics[width=1\textwidth]{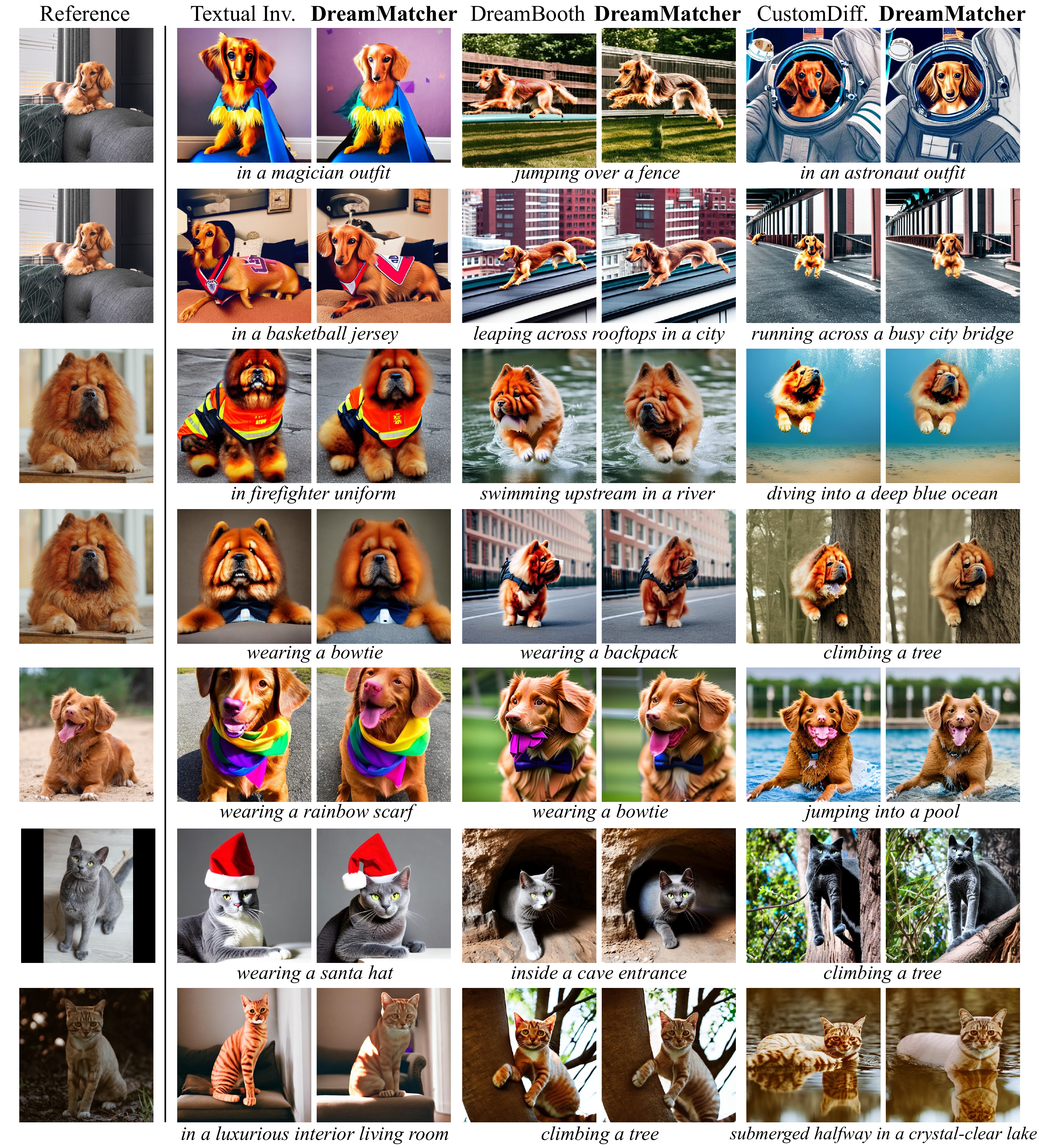} 
    \end{center}
    \vspace{-15pt}
    \caption{\textbf{Qualitative comparision with baselines for live objects: } We compare DreamMatcher with three different baselines, Textual Inversion~\cite{gal2022image}, DreamBooth~\cite{ruiz2023dreambooth}, and CustomDiffusion~\cite{kumari2023multi}.
}
    \vspace{-10pt}
    \label{supp:qual_with_base}
\end{figure*}

\begin{figure*}[t]
    \begin{center}
    \includegraphics[width=1\textwidth]{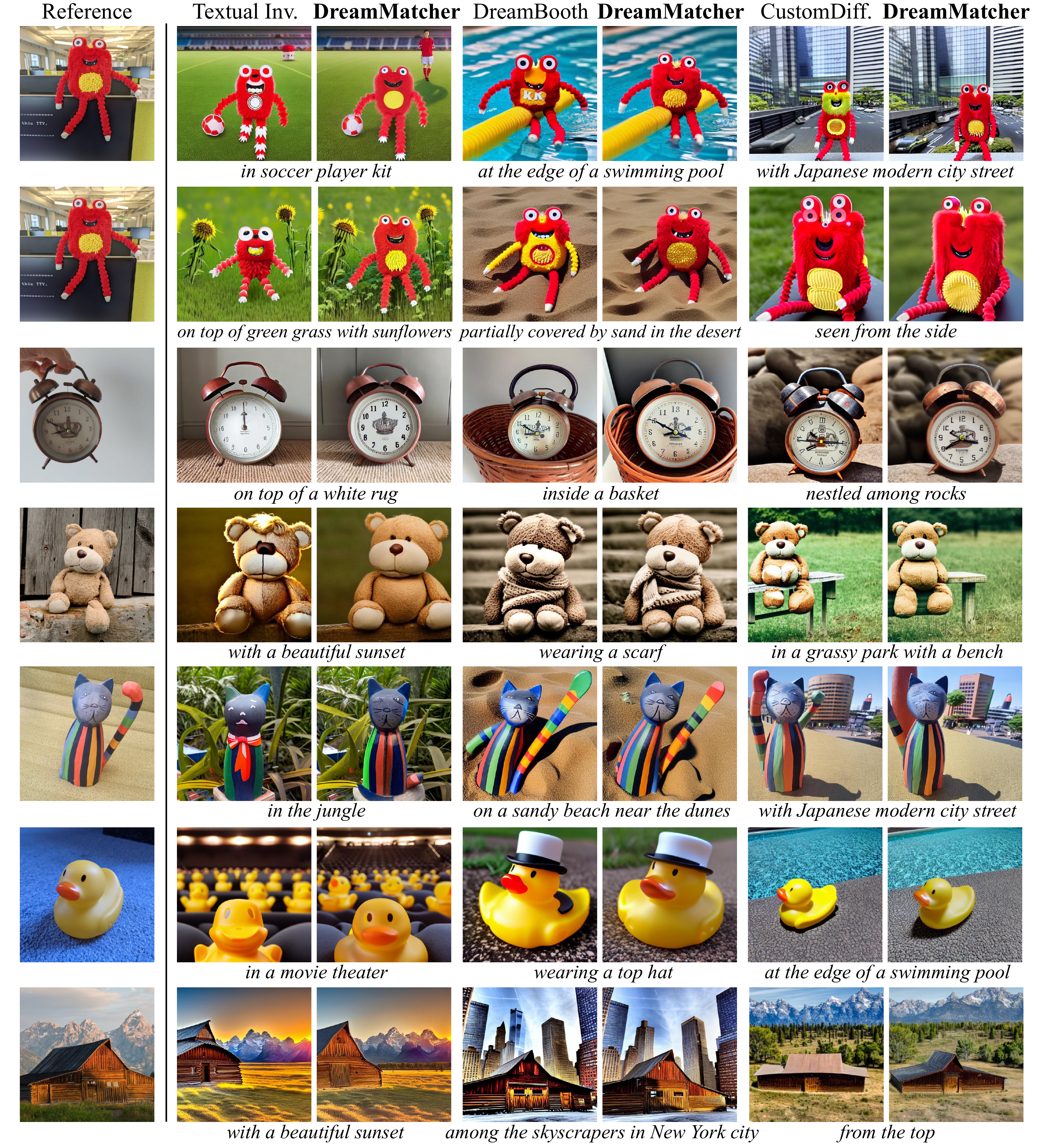} 
    \end{center}
    \vspace{-15pt}
    \caption{\textbf{Qualitative comparision with baselines for non-live objects: } We compare DreamMatcher with three different baselines, Textual Inversion~\cite{gal2022image}, DreamBooth~\cite{ruiz2023dreambooth}, and CustomDiffusion~\cite{kumari2023multi}.
}
    \vspace{-10pt}
    \label{supp:qual_with_base2}
\end{figure*}

We present more qualitative results comparing with baselines, Textual Inversion~\cite{gal2022image}, DreamBooth~\cite{ruiz2023dreambooth}, and CustomDiffusion~\cite{kumari2023multi} in Figure~\ref{supp:qual_with_base} and~\ref{supp:qual_with_base2}. 

\subsection{Comparison with Previous Works}
\label{supp:comp_sota}
\begin{figure*}[t]
    \begin{center}
    \includegraphics[width=1\textwidth]{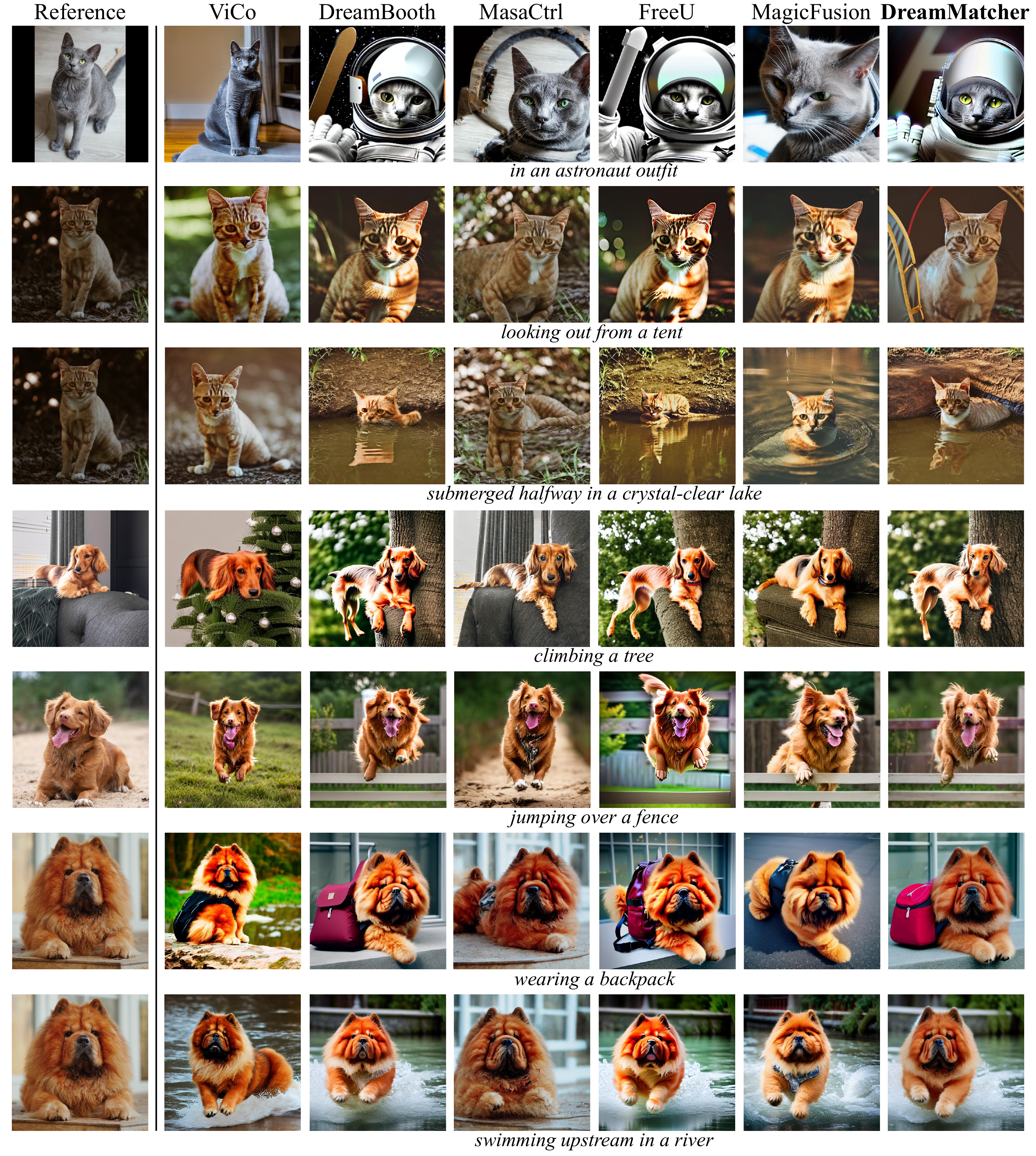} 
    \end{center}
    \vspace{-15pt}
    \caption{\textbf{Qualitative comparison with previous works~\cite{hao2023vico, ruiz2023dreambooth, cao2023masactrl, si2023freeu, zhao2023magicfusion} for live objects:} For this comparison, DreamBooth~\cite{ruiz2023dreambooth} is used as the baseline for MasaCtrl~\cite{cao2023masactrl}, FreeU~\cite{si2023freeu}, MagicFusion~\cite{zhao2023magicfusion}, and DreamMatcher.}
    \vspace{-10pt}
    \label{supp:qual_with_sota}
\end{figure*}

\begin{figure*}[t]
    \begin{center}
    \includegraphics[width=1\textwidth]{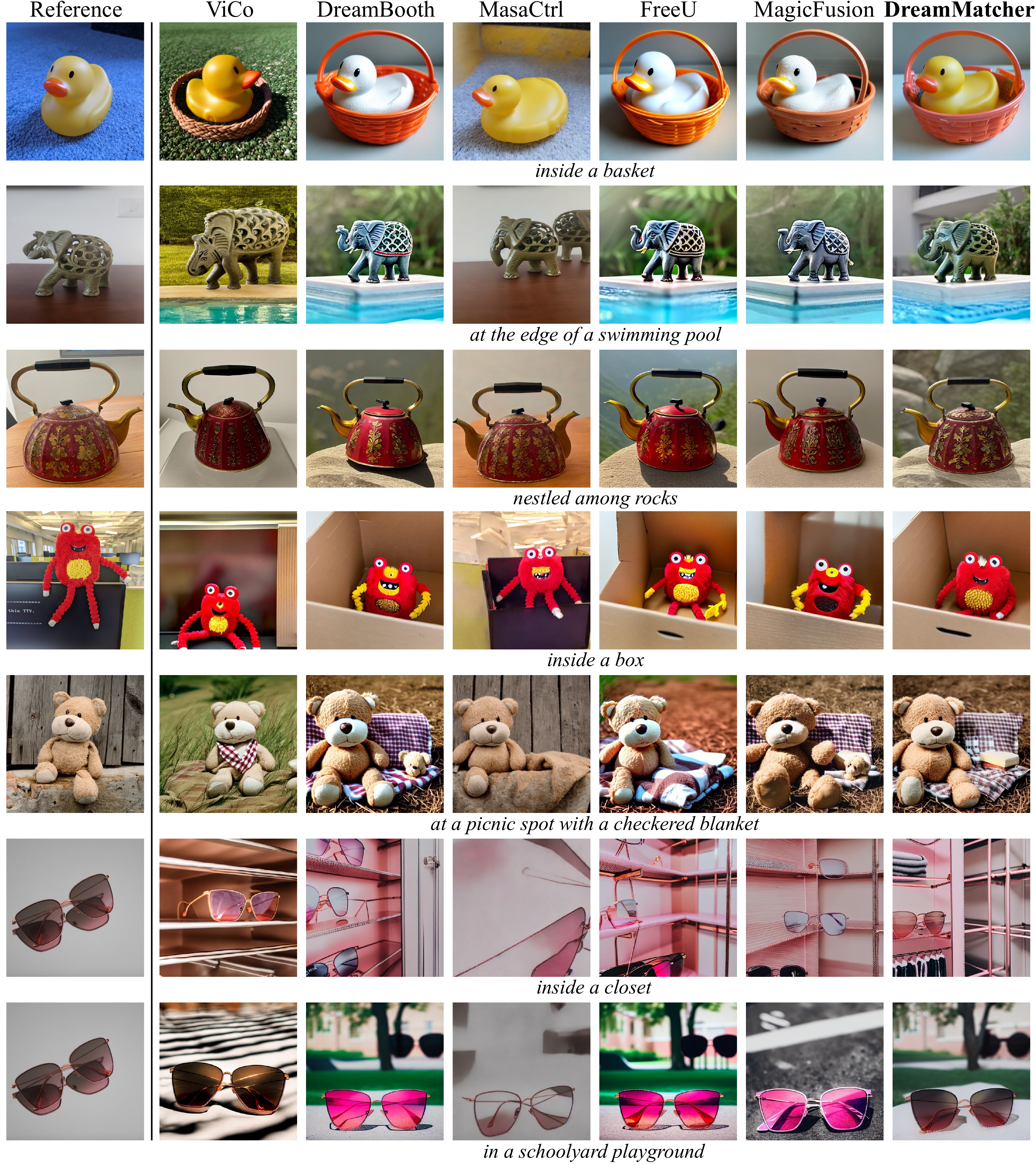} 
    \end{center}
    \vspace{-15pt}
    \caption{\textbf{Qualitative comparison with previous works~\cite{hao2023vico, ruiz2023dreambooth, cao2023masactrl, si2023freeu, zhao2023magicfusion} for non-live objects:} For this comparison, DreamBooth~\cite{ruiz2023dreambooth} is used as the baseline for MasaCtrl~\cite{cao2023masactrl}, FreeU~\cite{si2023freeu}, MagicFusion~\cite{zhao2023magicfusion}, and DreamMatcher.}
    \vspace{-10pt}
    \label{supp:qual_with_sota2}
\end{figure*}

We provide more qualitative results in Figure~\ref{supp:qual_with_sota} and~\ref{supp:qual_with_sota2} by comparing DreamMatcher with the optimization-based method ViCo~\cite{hao2023vico} and tuning-free methods MasaCtrl~\cite{cao2023masactrl}, FreeU~\cite{si2023freeu}, and MagicFusion~\cite{zhao2023magicfusion}. 

\begin{figure}[t]
    \centering
    \includegraphics[width=0.45\textwidth]{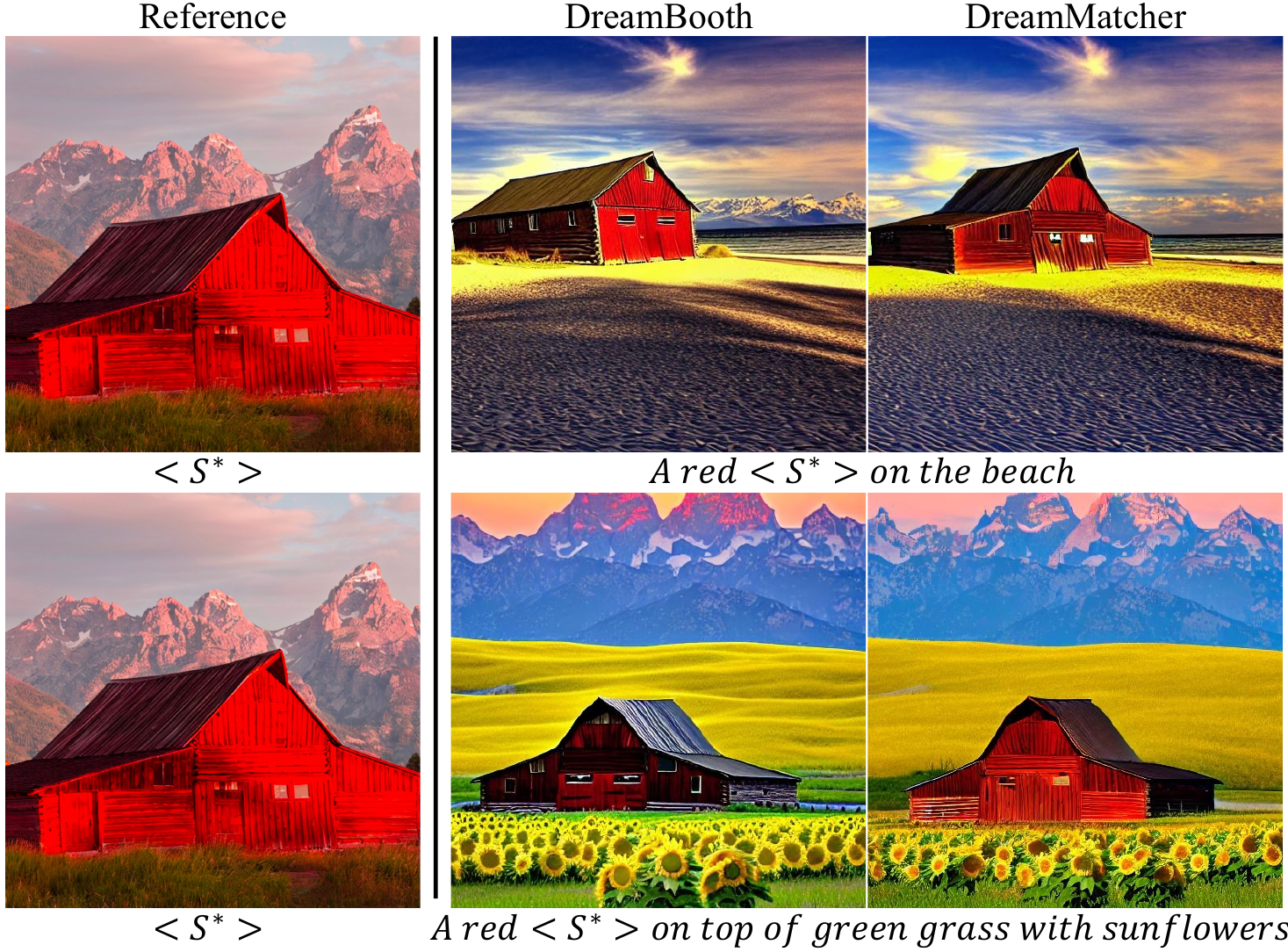}
    \vspace{-10pt}
    \caption{\textbf{Integrating an image editing technique:} From left to right: the edited reference image by instructPix2Pix~\cite{brooks2023instructpix2pix}, the target image generated by the baseline, and the target image generated by DreamMatcher with the edited reference image. DreamMatcher can generate novel subject images by aligning the modified appearance with diverse target layouts.}
    \label{supp:qual_editing}
    \vspace{-5pt}
\end{figure}

\begin{figure}[t]
    \centering
    \includegraphics[width=0.45\textwidth]{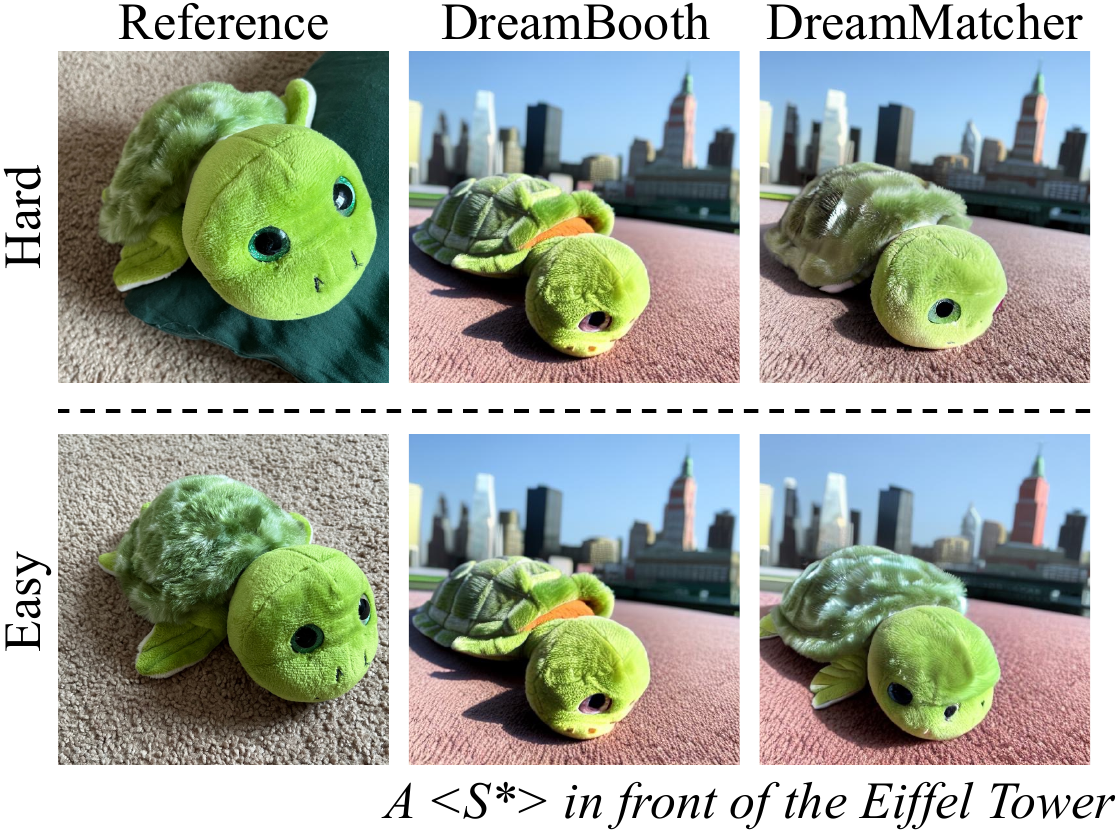}
    \vspace{-10pt}
    \caption{\textbf{Impact of reference selection on personalization:} The top row presents results using a reference image that is difficult to match, while the bottom row shows results using a reference image that is relatively easier to match. The latter, containing sufficient visual attributes of the subject, leads to improved personalized results. This indicates that appropriate reference selection can enhance personalization fidelity.}
    \label{supp:qual_reference}
    \vspace{-10pt}
\end{figure}

\section{Limitation}
\label{sec:limitation}
\paragrapht{Stylization.} DreamMatcher may ignore stylization prompts such as \textit{A red $\langle S^* \rangle$} or \textit{A shiny $\langle S^* \rangle$}, which do not appear in the reference images, as the model is designed to accurately inject the appearance from the reference. However, as shown in Figure~\ref{supp:qual_editing}, combining off-the-shelf editing techniques~\cite{brooks2023instructpix2pix, hertz2022prompt, tumanyan2023plug} with DreamMatcher is highly effective in scenarios requiring both stylization and place alteration, such as \textit{A red $\langle S^* \rangle$ on the beach}. Specifically, we initially edit the reference image with existing editing methods to reflect the stylization prompt \textit{red $\langle S^* \rangle$}, and then DreamMatcher generates novel scenes using this edited image. Our future work will focus on incorporating stylization techniques~\cite{brooks2023instructpix2pix, hertz2022prompt, tumanyan2023plug} into our framework directly, enabling the model to manipulate the reference appearance when the target prompt includes stylization.

\paragrapht{Extreme Matching Case.} In Appendix~\ref{supp:reference_select}, we demonstrate that our proposed method exhibits robust performance with randomly selected reference images. However, as depicted in Figure~\ref{supp:qual_reference}, using a reference image that is relatively challenging to match may not significantly improve the target image due to a lack of confidently matched appearances. This indicates even if our method is robust in reference selection, a better reference image which contains rich visual attributes of the subject will be beneficial for performance. Our future studies will focus on automating the selection of reference images or integrating multiple reference images jointly.

\begin{algorithm*}[t]
\caption{Pseudo-Code for Appearance Matching Self-Attention, PyTorch-like}
\label{supp:pseudo1}
\definecolor{codeblue}{rgb}{0.25,0.5,0.5}
\definecolor{codekw}{rgb}{0.85, 0.18, 0.50}
\lstset{
 backgroundcolor=\color{white},
 basicstyle=\fontsize{7.5pt}{7.5pt}\ttfamily\selectfont,
 columns=fullflexible,
 breaklines=true,
 captionpos=b,
 commentstyle=\fontsize{7.5pt}{7.5pt}\color{codeblue},
 keywordstyle=\fontsize{7.5pt}{7.5pt}\color{codekw},
}
\begin{lstlisting}[language=python]
def AMA(self, pca_feats, q_tgt, q_ref, k_tgt, k_ref, v_tgt, v_ref, mask_tgt, cc_thres, num_heads, **kwargs):

    # Initialize dimensions and rearrange inputs.
    B, H, W = init_dimensions(q_tgt, num_heads)
    q_tgt, q_ref, k_tgt, k_ref, v_tgt, v_ref = rearrange_inputs(q_tgt, q_ref, k_tgt, k_ref, v_tgt, v_ref, num_heads, H, W)
    
    # Perform feature interpolation and rearrangement.
    src_feat, trg_feat = interpolate_and_rearrange(pca_feats, H)
    src_feat, trg_feat = l2_norm(src_feat, trg_feat)
    
    # Compute similarity.
    sim = compute_similarity(trg_feat, src_feat)
  
    # Calculate forward and backward similarities and flows.
    sim_backward = rearrange(sim, ``b (Ht Wt) (Hs Ws) -> b (Hs Ws) Ht Wt'')
    sim_forward = rearrange(sim, ``b (Ht Wt) (Hs Ws) -> b (Ht Wt) Hs Ws'')
    
    flow_tgt_to_ref, flow_ref_to_tgt = compute_flows_with_argmax(sim_backward, sim_forward)

    # Compute cycle consistency error and confidence.
    cc_error = compute_cycle_consistency(flow_tgt_to_ref, flow_ref_to_tgt)
    fg_ratio = mask_tgt.sum() / (H * W)
    confidence = (cc_error < cc_thres * H * fg_ratio)

    # Warp value and apply semantic-consistent mask.
    warped_v = warp(v_ref, flow_tgt_to_ref)
    warped_v = warped_v * confidence + v_tgt * (1 - confidence)
    warped_v = warped_v * mask_tgt + v_tgt * (1 - mask_tgt)

    # Perform self-attention.
    aff = compute_affinity(q_tgt, k_tgt)
    attn = aff.softmax(-1)
    out = compute_output(attn, warped_v)

    return out
\end{lstlisting}

\end{algorithm*}
\begin{algorithm*}[t]
\caption{Pseudo-Code for Semantic Matching Guidance, PyTorch-like}
\label{supp:pseudo2}
\definecolor{codeblue}{rgb}{0.25,0.5,0.5}
\definecolor{codekw}{rgb}{0.85, 0.18, 0.50}
\lstset{
 backgroundcolor=\color{white},
 basicstyle=\fontsize{7.5pt}{7.5pt}\ttfamily\selectfont,
 columns=fullflexible,
 breaklines=true,
 captionpos=b,
 commentstyle=\fontsize{7.5pt}{7.5pt}\color{codeblue},
 keywordstyle=\fontsize{7.5pt}{7.5pt}\color{codekw},
}
\begin{lstlisting}[language=python]

for i, t in enumerate(tqdm(self.scheduler.timesteps)):

    # Define model inputs.
    latents = combine_latents(latents_ref, latents_tgt)
    
    # Enable gradient computation for matching guidance.
    enable_gradients(latents)

    # Sampling and feature extraction from the U-Net.
    noise_pred, feats = self.unet(latents, t, text_embeddings)

    # Interpolation and concatenating features from different layers.
    src_feat_uncond, tgt_feat_uncond, src_feat_cond, tgt_feat_cond = interpolate_and_concat(feats)

    # Perform PCA and normalize the features.
    pca_feats = perform_pca_and_normalize(src_feat_uncond, tgt_feat_uncond, src_feat_cond, tgt_feat_cond)

    # Apply semantic matching guidance if required.
    if matching_guidance and (i in self.mg_step_idx):
        _, pred_z0_tgt = self.step(noise_pred, t, latents)
        pred_z0_src = image_to_latent(src_img)
        uncond_grad, cond_grad = compute_gradients(pred_z0_tgt, pred_z0_src, t, pca_feats)
        
        alpha_prod_t = self.scheduler.alphas_cumprod[t]
        beta_prod_t = 1 - alpha_prod_t
        noise_pred[1] -= grad_weight * beta_prod_t**0.5 * uncond_grad
        noise_pred[3] -= grad_weight * beta_prod_t**0.5 * cond_grad

    # Apply classifier-free guidance for enhanced generation.
    if guidance_scale > 1.0:
        noise_pred = classifier_free_guidance(noise_pred, guidance_scale)

    # Step from z_t to z_t-1.
    latents = self.step(noise_pred, t, latents)
    
\end{lstlisting}

\end{algorithm*}

\clearpage
\newpage


\end{document}